\documentclass[a4paper]{ustthesis}
\usepackage[square,numbers]{natbib}
\usepackage[T1]{fontenc}
\usepackage{algpseudocode}
\usepackage{longtable}
\usepackage{url}
\usepackage{multirow}
\usepackage{graphics}
\usepackage{graphicx}
\usepackage{amsmath}
\usepackage{cases}
\usepackage{colortbl}
\usepackage{xcolor}
\usepackage{times}
\usepackage{array}
\usepackage{hyperref}
\usepackage{pdflscape}

\usepackage{diagbox}
\usepackage[T1]{fontenc}    
\usepackage{booktabs}       
\usepackage{amsfonts}       
\usepackage{nicefrac}       
\usepackage{microtype}      
\usepackage{xcolor}         
\usepackage{times}
\usepackage{epsfig}
\usepackage{graphicx}
\usepackage{amsmath}
\usepackage{amssymb}

\usepackage{amsmath}
\usepackage{amsfonts}
\usepackage{dsfont}
\usepackage{multirow}
\usepackage{adjustbox} 
\usepackage{wrapfig}
\usepackage{threeparttable}    

\usepackage[ruled,linesnumbered]{algorithm2e}
\usepackage{subfigure}
\makeatletter

\newcommand{\Rmnum}[1]{\expandafter\@slowromancap\romannumeral #1@}

\DeclareMathAlphabet\mathbfcal{OMS}{cmsy}{b}{n}

\newcommand{\eat}[1]{}
\ifodd 1

\newcommand{\TODO}[1]{{\color{red}TODO:{#1}}}

\else

\newcommand{\TODO}[1]{}

\fi

\title{Financial Knowledge Large Language Model}  
\author{Cehao YANG, Chengjin XU, Yiyan QI}     
\subject{} 
\department{}       
\advisor{}     
\member{}

\depthead{}     

\begin{document}

\newpage

\abstract
Artificial intelligence is making significant strides in the finance industry, revolutionizing how data is processed and interpreted. Among these technologies, large language models (LLMs) have demonstrated substantial potential to transform financial services by automating complex tasks, enhancing customer service, and providing detailed financial analysis. However, the application of LLMs in finance isn't without its challenges. One of the primary concerns is their tendency to generate hallucinations or spurious outputs, which can be particularly problematic in the finance sector where accuracy and reliability are paramount. Additionally, the cost and logistical challenges associated with regularly updating these models to reflect the latest financial regulations, market conditions, and economic data can be considerable. Moreover, there remains a significant challenge in how to effectively enhance LLMs with factual knowledge. Integrating robust, up-to-date factual knowledge into LLMs is critical to ensure that they can be reliably used in financial decision-making processes.

Firstly, we introduce IDEA-FinBench, an evaluation benchmark specifically tailored for assessing financial knowledge in large language models (LLMs). This benchmark utilizes questions from two globally respected and authoritative financial professional exams, aimimg to comprehensively evaluate the capability of LLMs to directly address exam questions pertinent to the finance sector. Secondly, we propose IDEA-FinKER, a Financial Knowledge Enhancement framework designed to facilitate the rapid adaptation of general LLMs to the financial domain, introducing a retrieval-based few-shot learning method for real-time context-level knowledge injection, and a set of high-quality financial knowledge instructions for fine-tuning any general LLM. Finally, we present IDEA-FinQA, a financial question-answering system powered by LLMs. This system is structured around a scheme of real-time knowledge injection and factual enhancement using external knowledge. IDEA-FinQA is comprised of three main modules: the data collector, the data querying module, and LLM-based agents tasked with specific functions.
\endabstract




\chapter{Introduction}
\label{chp_intro}
\section{Overview of AI in Finance}

Artificial intelligence (AI), particularly Natural Language Processing (NLP) technologies, has significantly transformed the finance industry. These technologies streamline complex tasks, automate customer services, and enhance decision-making processes. Large language models (LLMs), such as GPT-4, represent a cutting edge in this evolution, offering extensive capabilities that can analyze and interpret vast amounts of unstructured financial data quickly and effectively.

Despite their potential, the practical application of LLMs in finance faces substantial challenges. One of the primary concerns is their tendency for "hallucination," where the model might generate plausible but factually incorrect information. This is particularly problematic in finance, where accuracy and reliability are paramount. Additionally, the high cost of continuously updating these models to keep pace with the rapidly changing financial landscape poses another significant hurdle.

Moreover, there is an ongoing challenge in enhancing LLMs with factual knowledge. In finance, the demand for precise and up-to-date information is critical, and the static nature of trained models often clashes with the dynamic financial environment. To address these issues, the development of hybrid models that combine the generative power of LLMs with real-time, verified data sources could be a potential solution. This integration would aim to leverage the strengths of AI in understanding and processing language while ensuring the accuracy and reliability required in financial applications.

\section{Research Problems}

\paragraph{Financial Knowledge Benchmark} Although benchmarks for evaluating large language models (LLMs) in content generation strive for comprehensiveness and perfection, they often fall short in specialized fields such as finance, particularly in evaluative capacities. This shortfall has left unresolved the speculation about whether current popular LLMs hold professional skills and knowledge reserves on par with human financial experts and whether they are capable of handling automation tasks in the finance industry effectively.

\paragraph{Enhancement with Financial Knowledge} The adaptation of LLMs to specific domains, like finance, presents significant challenges. Attempts to enhance foundational models through further pre-training and fine-tuning with financial texts and instructional datasets have not led to the anticipated improvements. In some cases, these efforts have even caused a decline in performance. This indicates that the strategies for integrating financial knowledge into LLMs—whether through in-context learning or supervised fine-tuning—need further development and exploration.

\paragraph{Retrieval Augmented LLM} LLMs are inherently constrained by the scope of their training data, which typically represents a snapshot of internet corpora up to a certain temporal and spatial point. While trainers can control the spatial aspect of data consolidation, the temporal limitations present significant challenges for LLM applications that require up-to-date information beyond the training data's cutoff. Additionally, updating the models through methods like secondary pre-training or supervised fine-tuning is challenging and costly due to the delicate nature of model parameters and the substantial resources required.

\section{Thesis Outline}

In this thesis, we aim at building a trustworthy LLM in finance area, which is enhanced by factual knowledge.

In Chapter \ref{chp_finkbench}, we introduce IDEA-FinBench, a novel benchmark for assessing financial knowledge in LLMs by leveraging questions from two internationally recognized and esteemed financial professional examinations. These questions are presented in both Chinese and English, employ four distinct question formats, and cover sixteen financial disciplines. This comprehensive array allows for an in-depth evaluation of LLMs' proficiency in directly responding to finance-related examination questions. Furthermore, IDEA-FinBench incorporates a modular evaluation suite that supports the integration of external datasets, offering flexibility in customizing evaluation methods and interfacing with various LLMs. This feature enhances the adaptability and scalability of the evaluation framework.

In Chapter \ref{chp_finker}, we present IDEA-FinKER, a Financial Knowledge Enhancement Framework, aimed at facilitating the swift adaptation of general LLMs to specialized financial contexts at a reduced cost, eliminating the need for extensive external pre-training. IDEA-FinKER is supported by a meticulously curated and extensive database of Chinese financial examination questions and features an embedding similarity retrieval system. It underpins the development of a retrieval-based few-shot learning approach, termed the soft-injecting paradigm, for real-time contextual knowledge enhancement. Moreover, IDEA-FinKER introduces a structured set of financial knowledge instructions for fine-tuning general LLMs, described as the hard-injecting paradigm. Empirical results indicate that IDEA-FinKER markedly improves the expert-level capabilities of LLMs in the financial domain, significantly boosting their performance on IDEA-FinBench, particularly concerning Chinese financial examination questions such as the CPA.

In Chapter \ref{chp_finqa}, we introduce IDEA-FinQA, a dynamic financial question-answering system powered by LLMs. IDEA-FinQA operates under a real-time knowledge injection and factual enhancement paradigm, utilizing external knowledge bases. The system consists of three primary modules: the data collector, which is tasked with gathering and amalgamating data from the financial domain through both online and offline methods and data storage solutions; the data querying module, which implements search functionalities using both traditional text-based and contemporary embedding-based indexing systems for various recall and ranking stages; and four specialized LLM-based agents comprising a query rewriter, intention detector, extractor and refiner, and a response generator. Each agent is designed to perform specific tasks within different prompts and contexts, thereby driving the effectiveness of the IDEA-FinQA system.

The main contributions of of this thesis are:

\begin{enumerate}
    \item A benchmarking tool that evaluates the financial knowledge of LLMs using questions from prestigious global financial exams in both Chinese and English across sixteen disciplines, featuring a modular suite for flexible customization and scalability.
    
    \item A framework that enhances the rapid adaptation of LLMs to the financial sector through a comprehensive database of financial exam questions, supporting both soft and hard knowledge injection paradigms, leading to significant performance improvements in domain-specific applications.
    
    \item A financial question-answering system driven by LLMs, utilizing real-time knowledge injection and supporting various data collection and querying methodologies, structured around four specialized LLM-based agents for optimized task-specific responses.
    
\end{enumerate}

\newpage


\chapter{Preliminaries and Background}
\label{chp_review}
In Chapter \ref{chp_intro}, we emphasized the formidable performance of Pre-trained Language Models (PLMs) and their exceptionally outstanding capabilities in Natural Language Generation (NLG) and Natural Language Understanding (NLU). Therefore, we first deconstruct the most popular language model architecture to date, the Transformer Model. Subsequently, we introduce PLMs focusing on the different combinations of attention mechanisms, Encoder or Decoder modules, and pre-training tasks. Following this, we discuss Large Language Models (LLMs), particularly those that adhere to scaling laws for increasing parameters and corpus size, which have led to the emergence of In-Context Learning (ICL) \cite{brown2020language} capabilities. The final two chapters are dedicated to the application of LLMs in the finance sector, including models trained with financial knowledge and corresponding benchmarks. Moreover, we shift our focus to Trustworthy LLMs, investigating whether LLMs possess the capability to generate content that can withstand fact-checking and verification. Finally, we summarize our discussion on background.

\section{Transformer}

\begin{figure}[htbp]
    \centering
    \includegraphics[width=0.4\linewidth]{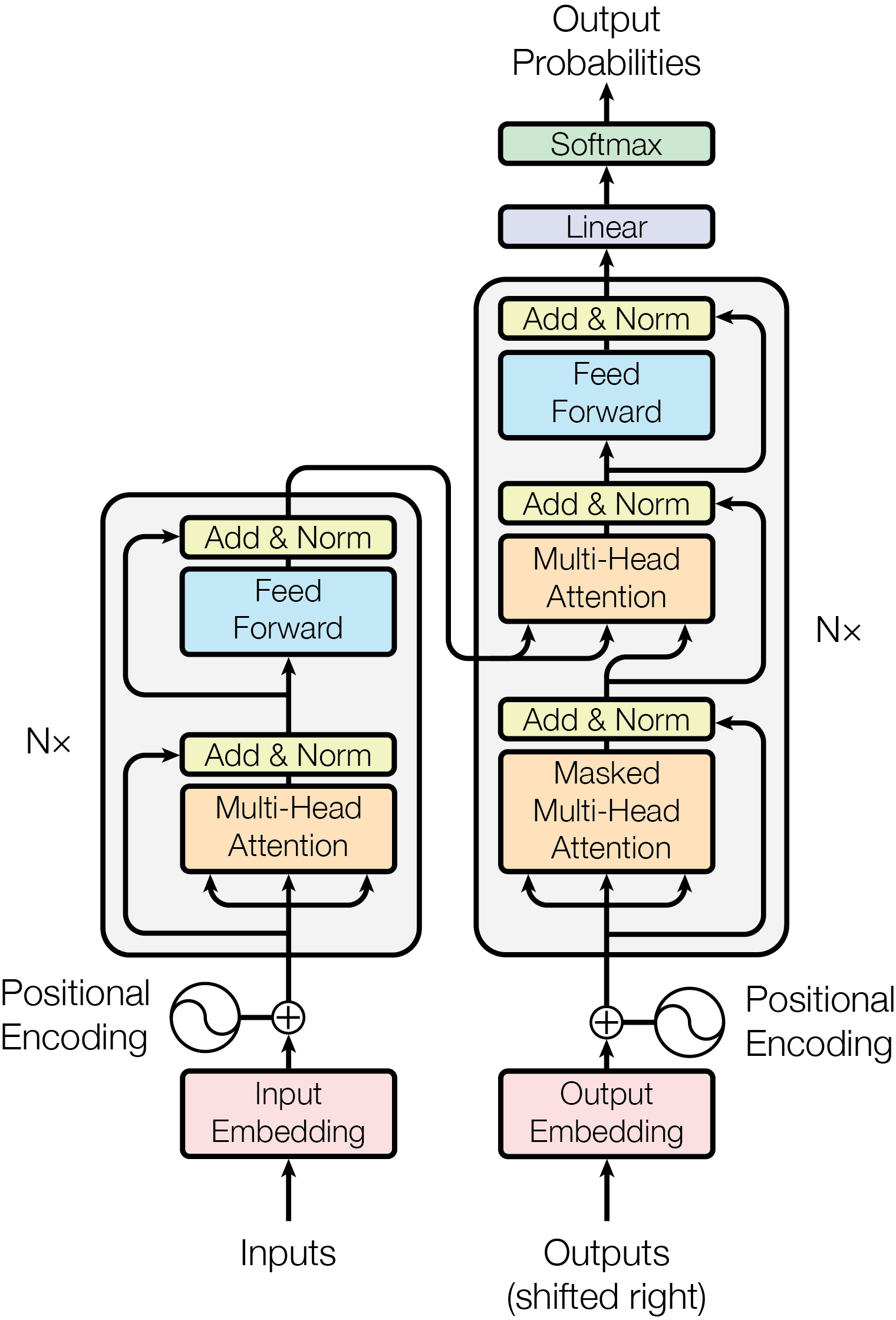}
    \caption{Architecture of Transformer Model \cite{vaswani2017attention}. The encoder (left) is stacked by multiple encoding layer, and the decoder (right) is stacked by multiple decoding layers.}
    \label{fig:arc_trans}
\end{figure}

 Natural Language Generation (NLG) tasks, such as machine translation, text summarization, question answering, text completion, and others, which not only demand a comprehensive understanding of natural languages, but also an additional downstream module is necessary to handle text generation. Sequence-to-sequence (seq2seq) \cite{sutskever2014sequence} learning has emerged as an exceedingly popular solution, defining the Encoder-Decoder as its foundational architecture. Here, the encoder part takes a sequence as input and maps it to a latent space to obtain a continuous representation, serving as the initial state for the decoder module, which generates words sequentially. As pioneering efforts, Recurrent Neural Networks (RNN) and their variations, such as LSTM (Long Short-Term Memory) \cite{hochreiter1997long} and GRU (Gated Recurrent Unit) \cite{chung2014empirical}, have been demonstrated to achieve tremendous success in NLG tasks, especially in the realm of machine translations. The attention mechanism was also initially introduced to seq2seq, designed to guide each word to adjust the weights of its attention across different words in the sequence through a scoring module \cite{bahdanau2014neural}. However, due to difficulties in parallelizing the model and limitations imposed by the hidden state that cap the network's ability to process long sequences of information and uncontrollable gradient variations during training, the Transformer model \cite{vaswani2017attention} made a striking entrance. Its unique self-attention mechanism not only introduces an exceptionally performant parallelism to the training and inference processes but also makes capturing long-range dependencies in sequences more feasible through positional encoding. Figure \ref{fig:arc_trans} shows an architecture of Transformer.

\section{Pre-trained Language Models}

As Transformer models continue to unlock their seemingly boundless potential for performance, the traditional paradigm of fully supervised learning is increasingly challenged by the pre-training and fine-tuning approach. Pre-trained Language Models (PLMs) adjust their vast number of parameters through unsupervised training on large-scale corpora, thereby incorporating the patterns and knowledge embedded within vast amounts of natural language, which are then applied to downstream tasks. By deconstructing the transformer model and selecting different pre-training tasks, PLMs are not confined to a singular Encoder-Decoder architecture. 

\subsection{Encoder-only Architecture}

\paragraph{BERT} BERT \cite{devlin2018bert} (Bidirectional Encoder Representations from Transformers) is fundamentally designed to comprehend sequences of language and map them into a semantic space, serving as the objective of its training regimen. At its core, BERT employs the Transformer's Encoder architecture, which it stacks in multiple layers to form a deep neural network. This architecture enables the model to capture complex syntactic and semantic relationships within text, as shown in Figure \ref{fig:arc_bert}. The groundbreaking aspect of BERT lies in its pre-training methodology, which utilizes a large corpus of unlabeled text to learn a rich representation of language. Through this unsupervised learning approach, BERT acquires a generalized understanding of language, which can then be fine-tuned for a wide array of downstream tasks. BERT's pre-training involves two primary tasks: 

\begin{figure}[htbp]
    \centering
\includegraphics[width=0.8\linewidth]{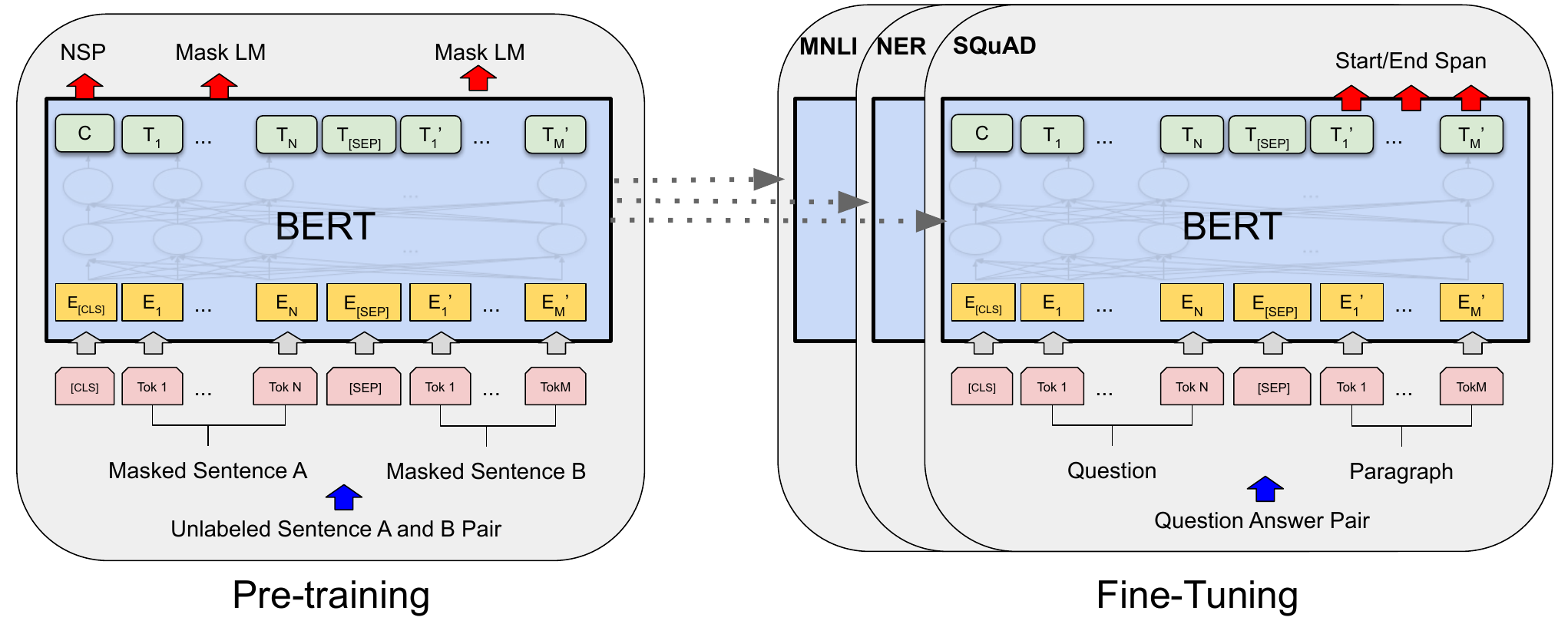}
    \caption{Architecture of BERT Model \cite{devlin2018bert}. It adopts unsupervised-learning during the pre-training stage (left), and adopts fine-tuning (right) to adapt the downstream tasks.}
    \label{fig:arc_bert}
\end{figure}

\begin{enumerate}

    \item Masked Language Modeling (MLM): a certain percentage of the input tokens are randomly masked, and the objective is for the model to predict the original identity of these masked tokens, based solely on their context. This task encourages the model to develop a deep understanding of language context and word relationships.

    \item Next Sentence Prediction (NSP): it aims to predict whether two given sentences logically follow each other. This is achieved by providing the model with pairs of sentences as input and training it to distinguish between pairs where the second sentence is a logical continuation of the first and pairs where it is not. 
    
\end{enumerate}

\subsection{Decoder-only Architecture}

\begin{figure}[htbp]
    \centering
    \includegraphics[width=0.9\linewidth]{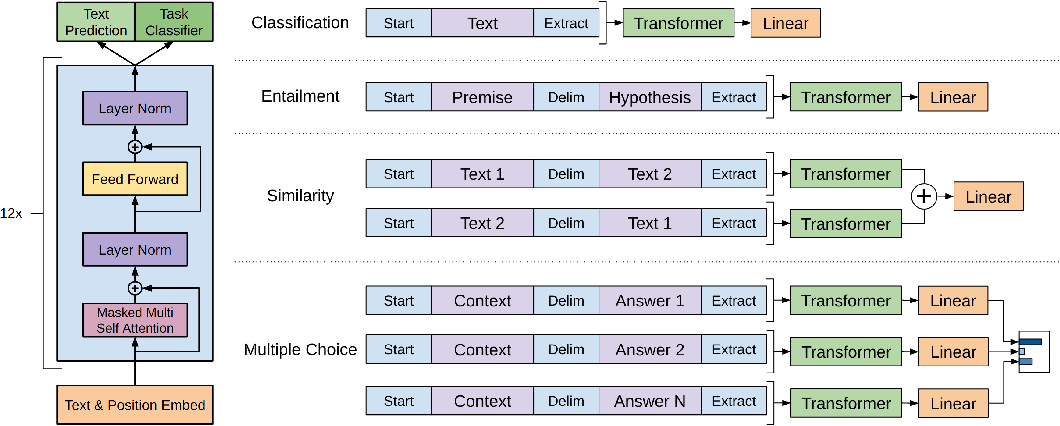}
    \caption{Architecture of GPT Model \cite{radford2018improving}. After pre-training through NSP tasks on a large unsupervised text corpus, it can be adapted to downstream tasks.}
    \label{fig:arc_gpt}
\end{figure}

\paragraph{GPT} GPT \cite{radford2018improving} (Generative Pre-trained Transformer) introduces a multi-layer Transformer decoder architecture, characterized by its unidirectional decoding process which is shown in Figure \ref{fig:arc_gpt}. This architecture adheres to the pre-training and fine-tuning paradigm, wherein the model undergoes unsupervised training on a large corpus through next token prediction (NTP) tasks, followed by fine-tuning on downstream datasets to tailor the model for specific tasks. The unsupervised NTP training objective under an unsupervised corpus is shown below:

Given an unsupervised text sequence \(T = \{t_1, t_2, \ldots, t_n\}\) as a prefix, the objective of the language model with parameters \(\theta\) is to maximize the likelihood of the sequence, which can be formulated as:
\begin{equation}
    \max_{\theta} \prod_{i=1}^{n} P(t_i | t_1, t_2, \ldots, t_{i-1}; \theta)
    \label{eq:ntp_1}
\end{equation}

This can also be expressed in terms of the log-likelihood to simplify computation:
\begin{equation}
    \max_{\theta} \sum_{i=1}^{n} \log P(t_i | t_1, t_2, \ldots, t_{i-1}; \theta)
    \label{eq:ntp_2}
\end{equation}

The parameters \(\theta\) are updated through gradient-based optimization methods to maximize the log-likelihood of the observed text sequence. As the scale of the training corpus and the model's parameters expanded, GPT-2 \cite{radford2019language} demonstrated its capability for unsupervised multitask learning, exhibiting the ability to perform specific tasks in a zero-shot setting without task-specific training. With the advent of GPT-3 \cite{brown2020language}, the potential of causal language models for in-context learning was fully realized. This involves crafting inputs that include a few example prompts as context, enabling the model to perform tasks without the need for any gradient updates, thereby avoiding the costs associated with extensive fine-tuning. Furthermore, the iterative development of the GPT series has empirically validated the scaling laws \cite{kaplan2020scaling}, indicating that model performance improves with increases in the scale of model parameters and training data.

\subsection{Encoder-Decoder Architecture}

\paragraph{T5} T5 \cite{raffel2020exploring} (Text-to-Text Transfer Transformer) adheres to the canonical Encoder-Decoder structure of the Transformer architecture. During its pre-training phase, T5 employs a BERT-like MLM \cite{devlin2018bert} task as its unsupervised learning strategy. In the fine-tuning phase, T5 redefines various NLP tasks within a uniform Text-to-Text paradigm, thereby allowing for a consistent training objective across multiple downstream tasks. This is achieved by appending specific instructions as prefixes to sub-tasks, enabling the model to distinguish among tasks during training and to maintain the same input-output format during inference.

\section{Alignment of LLMs}

\begin{figure}[htbp]
    \centering
    \includegraphics[width=0.9\linewidth]{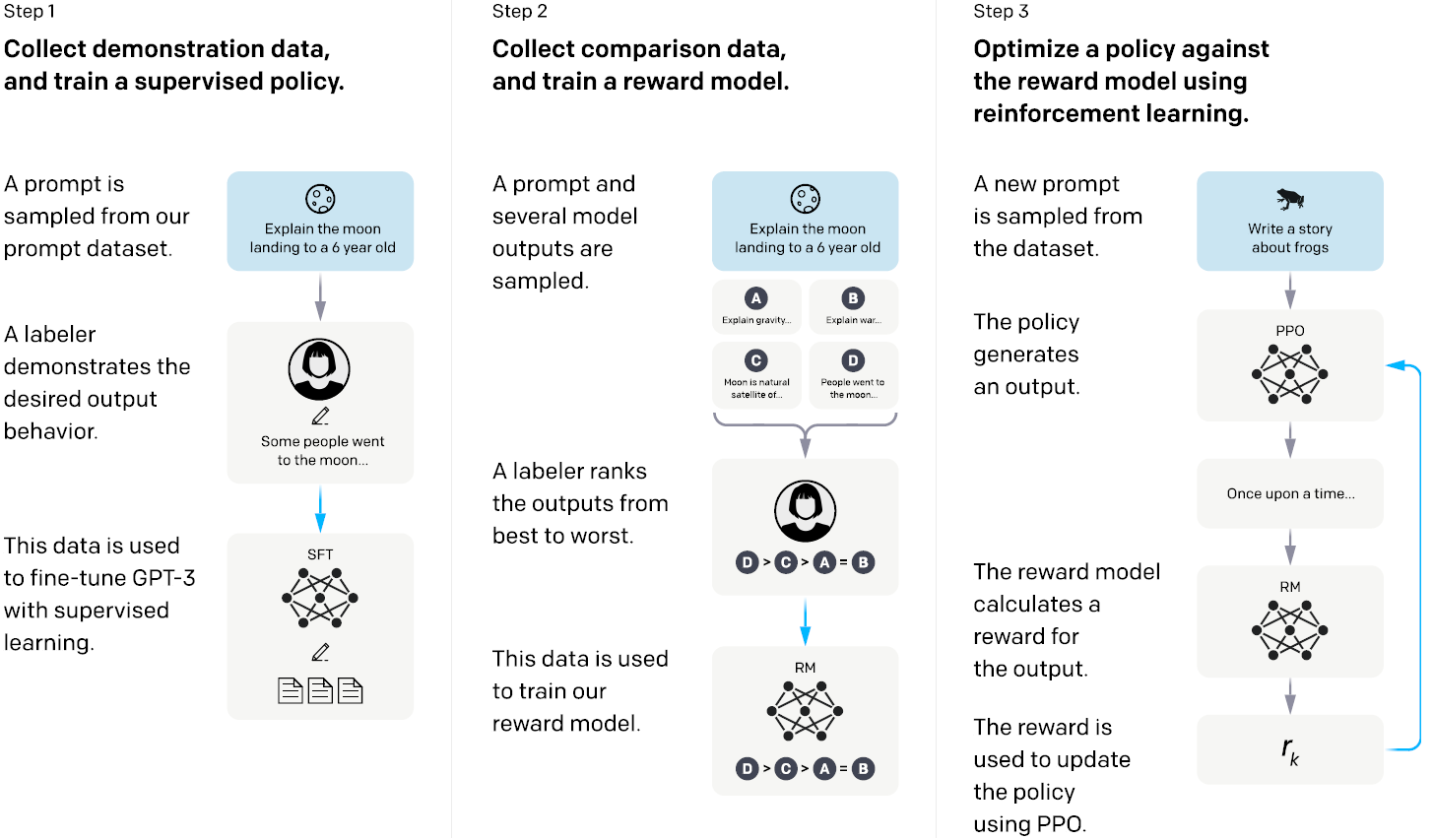}
    \caption{The alignment of InstructGPT \cite{ouyang2022training}. The first step one the left is supervised fine-tuning. The second step in the middle is the reward model training. The third step on the right is the reinforcement learning.}
    \label{fig:instructgpt}
\end{figure}

Alignment in the context of LLMs refers to the process and the goal of making these models understand and follow human intentions, ethics, and instructions as closely as possible \cite{gabriel2020artificial, kenton2021alignment, askell2021general}. The concept of alignment is centered around ensuring that the outputs of LLMs are not only relevant and informative but also adhere to the moral and social norms that govern human discourse. This involves training models to avoid generating harmful, biased, or misleading content while enhancing their ability to produce responses that are truthful, appropriate, and aligned with the explicit instructions or goals set by their users \cite{weidinger2021ethical}. Some pioneering work \cite{ouyang2022training} has already attempted to enhance the alignment of language models with human intentions by utilizing a two-step process that combines Supervised Fine-Tuning (SFT) with Reinforcement Learning from Human Feedback (RLHF) \cite{christiano2017deep}, leveraging human-written prompts and feedback to finely adjust its responses, which is shown in Figure \ref{fig:instructgpt}.

\section{Language Models in Finance}\label{llm_fin}

\subsection{Language Models for Financial NLU Tasks}

The application of NLP techniques to financial scenarios has evolved significantly over time. Initially, early works focused primarily on NLU tasks within the financial domain. This included efforts in financial entity recognition \cite{wang2014financial}, sentiment analysis \cite{chan2017sentiment}, financial event extraction \cite{yang2018dcfee}, financial question answering \cite{maia201818}, and text summarization \cite{abdaljalil2021exploration}. These foundational tasks aimed to enhance the comprehension and processing of financial texts, which are often complex and laden with domain-specific terminology. 

With the advent of BERT \cite{devlin2018bert}, a shift occurred towards leveraging PLMs for improved performance in financial NLP tasks. Among these advancements, FinBERT \cite{araci2019finbert,yang2020finbert,liu2021finbert} stands out as a prominent example. FinBERT usually builds upon the BERT model by selectively applying incremental pre-training on a large corpus of financial texts and fine-tuning. This domain adaptation process is specifically tailored to enhance the model's performance on financial-specific downstream tasks. BBT-FinT5 \cite{lu2023bbt}, which is pre-trained and fine-tuned on the T5 architecture based on a large-scale financial corpus, also verifies the potential of PLM to unleash powerful performance in specific areas.

\subsection{Generative Financial LLMs}

The emergence and global popularity of LLM-based dialogue systems, such as ChatGPT \cite{OpenAI2022ChatGPT} and GPT-4 \cite{openai2023gpt4}, have further advanced the field. These models have showcased remarkable capabilities in instruction following and zero-shot learning across a wide range of applications. The auto-regressive architecture of generative LLMs, in particular, has drawn attention for its potential in breaking new ground in NLG within the financial domain. This has sparked expectations for achieving breakthroughs in tasks long considered exclusive to human expertise in finance, such as investment management, risk modeling, and customer advisory services \cite{li2023large}. The prospect of achieving Artificial General Intelligence (AGI) in finance, capable of autonomously performing a broad spectrum of financial operations, is now seen as more feasible.

\paragraph{Early works of Financial LLMs} The early work on adapting LLMs for the financial domain served as a pioneering effort \cite{wu2023bloomberggpt,zhang2023xuanyuan}, offering valuable training pipelines to the public, especially in scenarios with limited foundational model options, such as BLOOM \cite{le2022bloom}. However, due to the sensitive nature of financial data and various corporate confidentiality policies, these leading efforts failed to contribute valuable and substantial financial corpora to the community. FinGPT \cite{yang2023fingpt, liu2023fingpt} aims to democratize access to financial data, building an open and transparent community for the development of financial language models. Subsequently, the open access of the decoder-only causal language model, LLaMA \cite{touvron2023llama,touvron2023llama2}, has been well received due to its range of model sizes from 6B to 70B parameters. This range allows for more accessible hardware support and reduced computational costs for incremental pre-training and fine-tuning, fostering the development of financial-domain-specific LLaMAs \cite{Cornucopia-LLaMA-Fin-Chinese,Todt_Bavest_FIN-LLAMA}.

\begin{figure}[htbp]
    \centering
\includegraphics[width=0.8\linewidth]{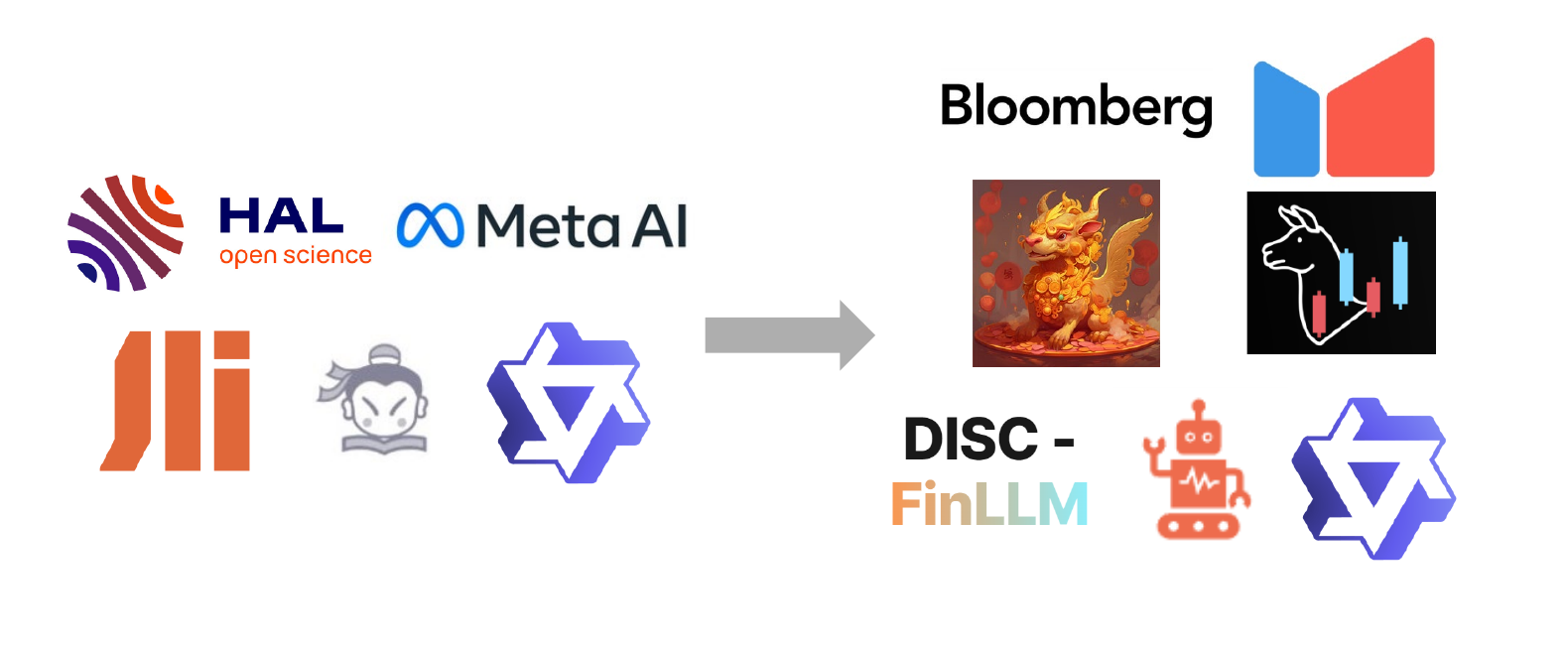}
    \caption{Base pre-trained language models (left) and financial-domain LLMs (right).}
    \label{fig:finllms}
\end{figure}

\paragraph{Chinese Financial LLMs}High-quality open-source projects for Financial LLM \cite{chen2023disc,li2023cfgpt,tongyi_finance_14b_chat} that perform external training on Chinese foundational language models \cite{yang2023baichuan, team2023internlm, bai2023qwen} also merit attention. These  projects selectively utilize industry corpora such as Chinese enterprise research reports, public company announcements, and financial news for incremental pre-training. They also employ knowledge-intensive downstream task instructions for fine-tuning, adapting to different financial expert roles, and may create a significant impact in the Chinese financial automation office scenario.

\subsection{Financial Benchmarks for Evaluation} Inspired by the popularity of generic benchmarks for PLMs \cite{wang2018glue, hendrycks2020measuring, srivastava2022beyond, zhong2023agieval, huang2024ceval}, benchmarks tailored to the financial domain have gradually emerged to accommodate the rapid development of domain-specific language models. Initial financial benchmarks were integrated with natural language understanding, but tailored to the financial sector, exemplified by FLUE \cite{shah2022flue}, which introduced an evaluation suite for downstream tasks including entity recognition, sentiment analysis, and headline classification. With the widespread adoption of LLMs-based dialogue systems, generative tasks have gained prominence in various assessments. For instance, FinQA and ConvFinQA \cite{chen2021finqa, chen2022convfinqa} focus on the model's ability to apply chain-of-thought reasoning and perform numerical reasoning within single-turn or multi-turn dialogues to solve problems. BBT-CFLEB \cite{lu2023bbt} incorporates a significant number of generative tasks, mixing them with understanding tasks. FinEval \cite{zhang2023fineval}, which draws on the design philosophy of C-Eval \cite{huang2024ceval}, addresses the examination of financial knowledge by including a vast array of professional certification exam questions to assess whether models meet the entry requirements of China's financial industry.

\section{Trustworthy LLMs}

The excellent performance of PLMs in knowledge-intensive tasks such as open-domain question answering and fact verification \cite{roberts2020much, yu2022generate} suggests their significant potential in supplanting traditional knowledge bases, and in offering high-quality logical reasoning as a complement to structured queries \cite{petroni2019language, heinzerling2020language, alkhamissi2022review}. Current state-of-the-art models like GPT-4 \cite{openai2023gpt4} have demonstrated prowess beyond the general public and even human experts in popular benchmarks for common sense question answering and reasoning \cite{hendrycks2020measuring, zellers2019hellaswag}. However, tasks involving factual knowledge often cast doubt on the reliability of these language models, especially due to their propensity for producing hallucinations \cite{ji2023survey, chang2023survey} or inaccuracies in time-sensitive \cite{hu2023large, schuster2021get} or domain-specific contexts, such as the medical, law, and the financial fields we have discussed in Section \ref{llm_fin}.

\begin{figure}[htbp]
    \centering
\includegraphics[width=0.8\linewidth]{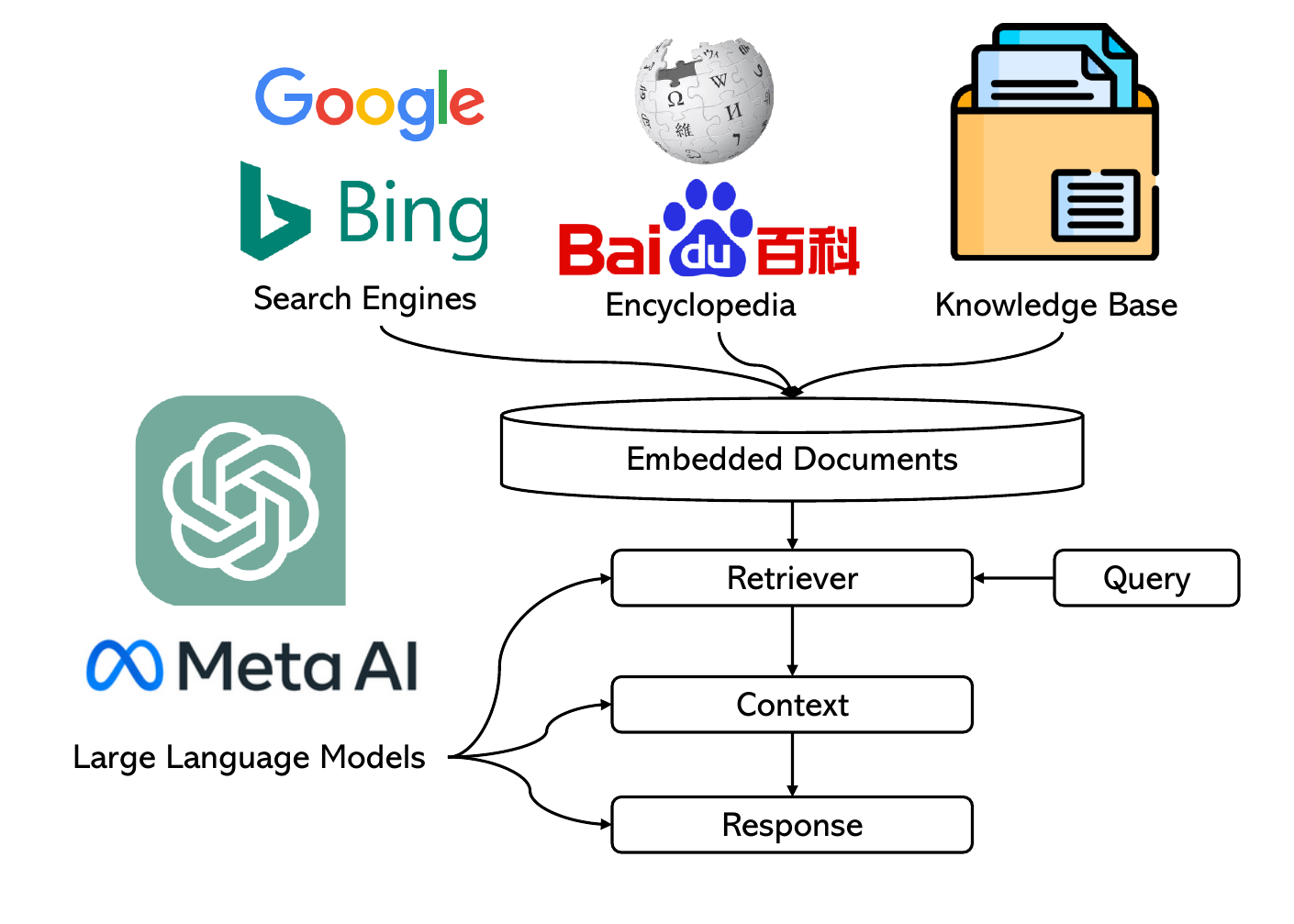}
    \caption{A sample pipeline of how to apply RAG with LLMs.}
    \label{fig:rag}
\end{figure}

\paragraph{Retrieval Augmented Generation} Retrieval Augmented Generation (RAG) has been demonstrated to effectively enhance the performance of language models on knowledge-intensive tasks by leveraging external knowledge \cite{lewis2020retrieval}. Observations of the emergent in-context learning \cite{brown2020language} capabilities of generative LLMs, coupled with the construction of chain-of-thought \cite{wei2022chain} prompts to guide more reliable reasoning pathways, position LLMs as possessing significant potential to act as retrievers within information systems \cite{zhu2023large}. Through predefined interfaces and the construction of corresponding queries, LLMs are endowed with the capability to access global knowledge, effectively mitigating the generation of hallucinations and the obsolescence of internalized parametric knowledge \cite{nakano2021webgpt, lazaridou2022internet, semnani2023wikichat, qin2023webcpm}. This has inspired the development of frameworks designed to inject external knowledge into LLMs, such as LlamaIndex \cite{liu2022llamaindex} and LangChain \cite{chase2022langchain}. BingChat \cite{Microsoft2023bing}, as a dialogue system enhanced by internet search engine capabilities, exemplifies the pioneering role of RAG at the application level. An overview of RAG is shown in Figure \ref{fig:rag}.

\paragraph{Benchmarks for Factual Knowledge} Recent evaluation benchmarks have emerged to assess the capability of LLMs to accommodate, master, and apply factual knowledge stored in their parameters. Some of these studies deliberately examine the veracity of the content generated by language models, particularly concerning common misconceptions, stereotypes, or logical fallacies prevalent in human society \cite{lin2021truthfulqa, yin2023large, li2023halueval}. This scrutiny challenges models that merely mimic human text without undergoing proper alignment to learn to acknowledge "Unknown" due to insufficient knowledge or "refuse to answer" when faced with factual inaccuracies in questions \cite{zhang2023r}. Additionally, evaluation suites that directly quantify the response quality of models using factual knowledge question-answer pairs constructed from authoritative knowledge encyclopedias, such as Wikipedia, as reliable sources of knowledge, have also been introduced \cite{hu2023large, vu2023freshllms, mallen2022not}. These efforts extend the design principles of traditional PLM tasks, such as fact-checking or fact verification \cite{thorne2018fever, aly2021feverous, jiang2020hover}, but exclusively retain the claim part as the input for LLMs.

\section{Summary}

In summary, we initially introduced the Transformer Model, the currently most prevalent deep neural network architecture. Particularly, we discussed PLMs based on the Transformer's encoder and decoder, covering some pioneering pre-training methodologies. As the scale of parameters expands and computational power increases, the emergence of intelligent language models capable of responding to prompts necessitates significant attention towards alignment. Returning to the core focus of our thesis, the LLMs in the financial sector, we reviewed open-source contributions and benchmarks. Another focal point is the trustworthiness of LLMs, especially those evaluated around factual knowledge and evaluation suites. In the next chapter, we will concentrate on constructing trustworthy LLMs within the financial sector to ensure fact-based questioning and reasoning.
\newpage

\chapter{IDEA-FinBench: Financial Knowledge Benchmark}
\label{chp_finkbench}
Despite the benchmarks for assessing the quality of content generation by large language models (LLMs) tending towards comprehensiveness and perfection in a general dimension, there remains a gap in specific fields, particularly in evaluative aspects of financial knowledge. This gap keeps the speculation that current popular LLMs possess professional skills and knowledge reserves comparable to human financial experts, and are competent for automation tasks in the finance industry, unresolved. Therefore, we aim to adopt a systematic research approach to explore how to comprehensively and objectively assess the ability of language models to grasp and apply financial knowledge for reasoning. Moreover, we focus on introducing LLMs enhanced with domain-specific financial knowledge.

We introduce IDEA-FinBench, an evaluation benchmark for financial knowledge in LLM, utilizing questions from two globally renowned and authoritative financial professional exams as the primary sources for assessment. The questions, encompassing both Chinese and English languages, four types of question formats, and spanning sixteen financial disciplines, are designed to evaluate LLMs' capabilities in directly addressing exam questions relevant to the finance sector comprehensively. Additionally, we provide a modular evaluation suite that can incorporate external datasets, allowing for flexible customization of evaluation modes and interfaces with various LLMs, thus offering adaptability and scalability to the evaluation framework. 

\section{Introduction}

Currently, generative pre-trained language models (GPLMs) have achieved exceptional results on a variety of standard benchmarks for natural language understanding and generation, reaching performances that rival the average human level \cite{openai2023gpt4}. This success is attributed to their extensive knowledge base and robust logical reasoning capabilities. Such evaluations often focus on skills traditionally believed to require advanced cognitive abilities and thought to be exclusive to humans, including logic, mathematics, coding, physics, and causal reasoning \cite{hendrycks2020measuring, zhong2023agieval, srivastava2022beyond}. These are considered prime choices for objectively assessing the potential intelligence level of large language models, which are increasingly approximating the capabilities of General Artificial Intelligence (AGI). However, in the financial sector, there remains a gap and a lack of comprehensive benchmarks. For instance, assessments of financial LLMs are still limited to traditional NLU tasks, such as entity recognition, sentiment analysis, and event extraction \cite{shah2022flue, lu2023bbt, lei2023cfbenchmark, chen2023disc}. These measures are inadequate for evaluating the models' abilities to perform high-level office tasks within the financial industry. Moreover, while some studies have recognized the potential for language models to perform intelligent tasks in finance, they have been hindered by a narrow focus on specific scenarios, causing a lack of generalizability, such as in numerical reasoning \cite{chen2021finqa, chen2022convfinqa} and investment decision-making \cite{liu2023fingpt}.

\begin{figure}[htbp]
    \centering
    \includegraphics[width=0.8\linewidth]{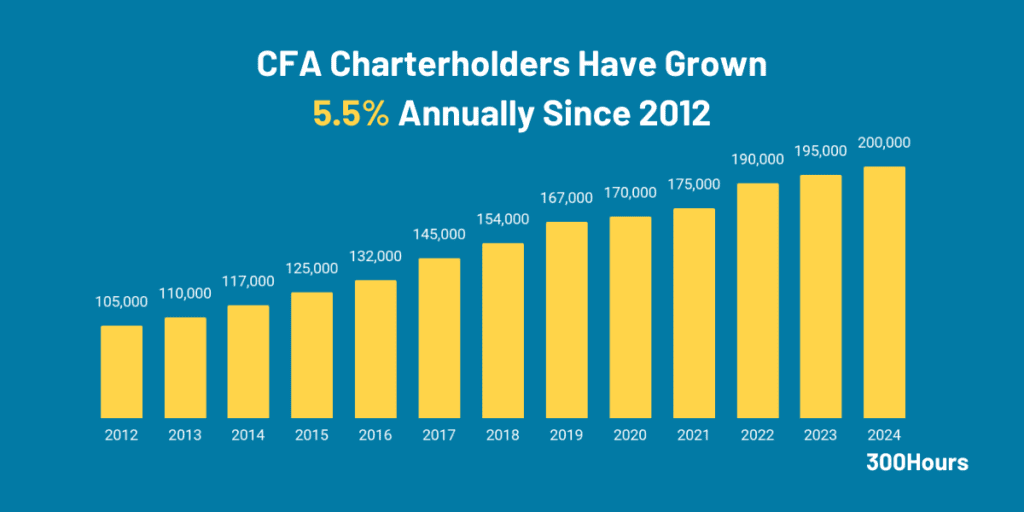}
    \caption{By February 2024, the global community of CFA charterholders has surpassed 190,000 individuals across over 160 countries, exhibiting a steady annual increase of 5.5\% throughout the decade from 2012 to 2022 \cite{300hours2024CFA}.}
    \label{fig:cfaholder}
\end{figure}

Wall Street, the epitome of top-tier financial firms, shows a strong preference for professionals holding the CFA certification, highlighting the importance of specialized training and a deep reserve of financial knowledge and problem-solving skills, which is shown in Figure \ref{fig:cfaholder}. In light of this, our benchmark evaluations aim to include the CFA certification to measure LLMs' capabilities in the financial sector. Despite the advanced abilities of current models like GPT-4, previous tests indicate they fall short of the CFA certification requirements \cite{callanan2023can}. Therefore, integrating these prestigious financial certification exams into our benchmarks is crucial to thoroughly assess LLMs' proficiency in finance.

We introduce IDEA-FinBench \footnote{The IDEA-FinBench data and evaluation code are available at https://github.com/IDEA-FinAI/IDEAFinBench.}, a benchmark that incorporates authoritative financial professional certification exams, including the Chinese Certified Public Accountant (CPA) and the international CFA exams. This benchmark spans both Chinese and English languages, four types of question formats, and sixteen financial subjects, ensuring a comprehensive assessment of LLMs' abilities to apply financial knowledge and solve real-world financial industry problems. Building upon the foundation of prior work on C-Eval \cite{huang2024ceval}, IDEA-FinBench offers the community a modular evaluation suite for assessing LLMs on multiple-choice question datasets. Our suite focuses on adaptability and scalability, supporting customization of reasoning modes and interface coupling for various LLMs. It includes a parallel mechanism for accelerated evaluation, a complete logging system, and cross-linguistic prompt settings.

We undertake a comprehensive integration of the various LLMs currently prevalent in the field, which encompasses both those that have open-sourced their model parameters and those that merely offer access to their model interfaces. By establishing a standardized and equitable inferencing framework, we have been able to derive a unified, quantifiable score. This score serves as a direct benchmark for assessing the LLMs' proficiency in grasping financial knowledge and their capacity to apply this understanding in reasoning and problem-solving contexts.

Our contributions can be summarized below:

\begin{enumerate}

    \item \textbf{Establishment of a Benchmark for Financial Knowledge Assessment in LLMs:} We introduced IDEA-FinBench, a fair benchmarking system for evaluating the intelligence level of LLMs regarding financial knowledge. Moreover, we have publicly disclosed the scores of popular LLMs on this benchmark, providing a transparent and standardized metric for comparison.
    
    \item \textbf{Development of an Adaptable and Scalable Evaluation Suite:} Our evaluation suite not only includes customization of reasoning modes and interface coupling for various LLMs, but also supports a parallel mechanism for accelerated evaluation, a comprehensive logging system, and cross-linguistic prompt settings.
    
\end{enumerate}

\section{Data Collection}

\subsection{Subject Selection}

In the construction of a robust and comprehensive financial knowledge benchmark, we incorporate Certified Public Accountant (CPA) and Chartered Financial Analyst (CFA) credentials as primary sources of high-caliber financial questions. 

\begin{figure}[htbp]
    \centering
\includegraphics[width=1.0\linewidth]{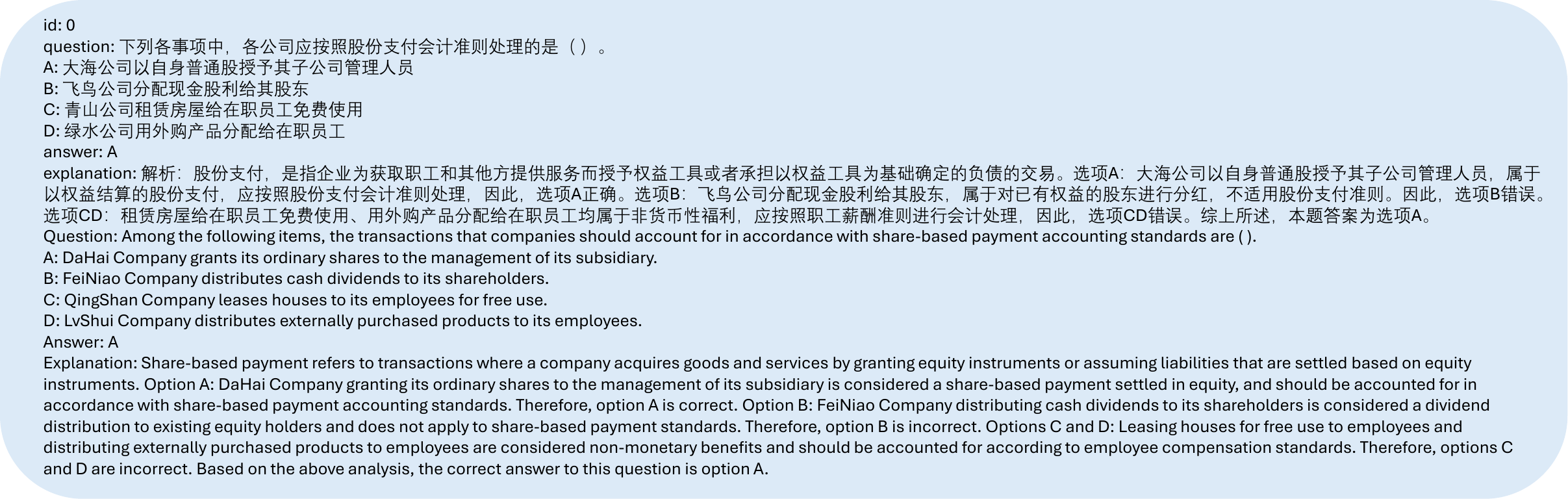}
    \caption{An example of CPA problem with only one answer.}
    \label{fig:cpa_one}
\end{figure}

\begin{figure}[htbp]
    \centering
\includegraphics[width=1.0\linewidth]{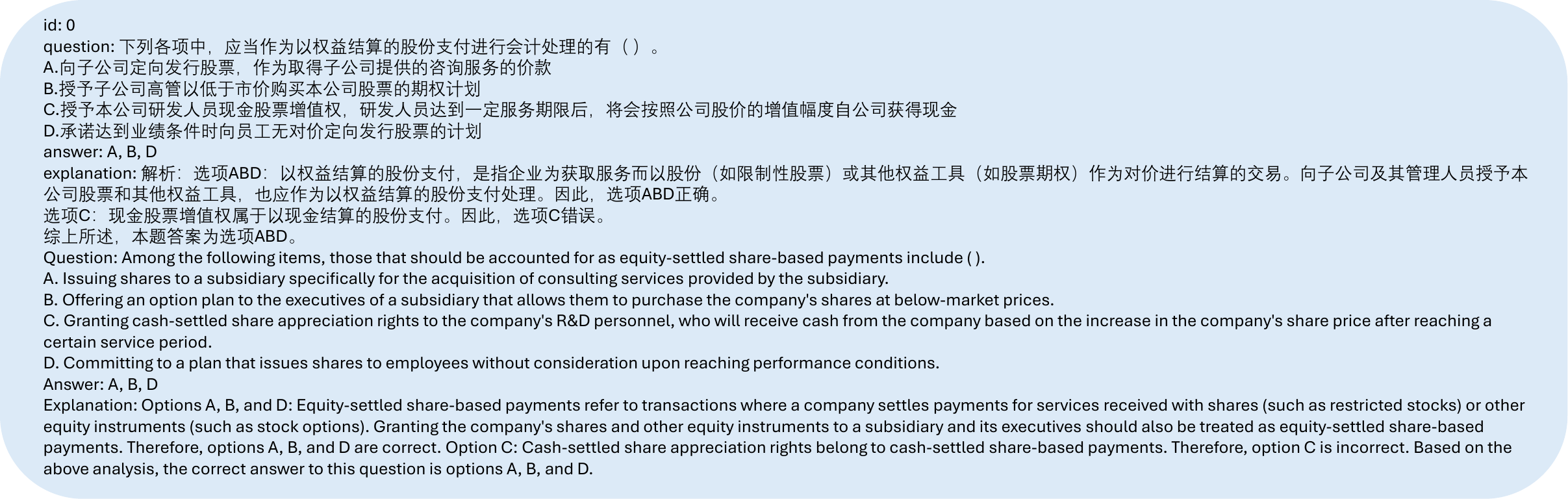}
    \caption{An example of CPA problem with more than one answer.}
    \label{fig:cpa_multi}
\end{figure}

\paragraph{Certified Public Accountant} The CPA dataset encompasses a broad spectrum of subjects including accounting, financial cost management, tax law, auditing, corporate strategy and risk management, and economic law. This selection aims to rigorously evaluate the capability of generic large models across a variety of fields such as accounting, auditing, and tax law. Analyzing the nature of the subjects, they can be categorized into computational and memorization domains. Computational subjects, which assess the model's ability to engage in financial logical reasoning, primarily include accounting, financial cost management, and tax law. Conversely, memorization subjects focus on the model's level of financial knowledge retention, covering auditing, corporate strategy and risk management, and economic law. Regarding the type of questions, the examination items are divided into questions with single answer and multiple answers. Examples problems are shown in Figure \ref{fig:cpa_one} and \ref{fig:cpa_multi}. Questions with multiple answers, in contrast to those with only one answer, challenge the generic large models to select more than one correct answer from several options, thereby testing their comprehensive analytical and judgment skills more rigorously. Furthermore, the assessment methodology for questions with multiple answers is inherently more complex.

\begin{figure}[htbp]
    \centering
\includegraphics[width=1\linewidth]{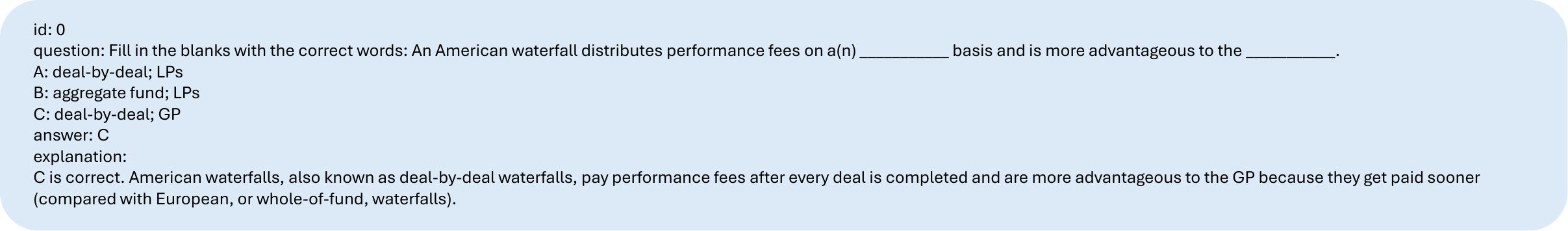}
    \caption{An example of CFA Level 1 problem.}
    \label{fig:cfa_l1}
\end{figure}

\begin{figure}[htbp]
    \centering
\includegraphics[width=1\linewidth]{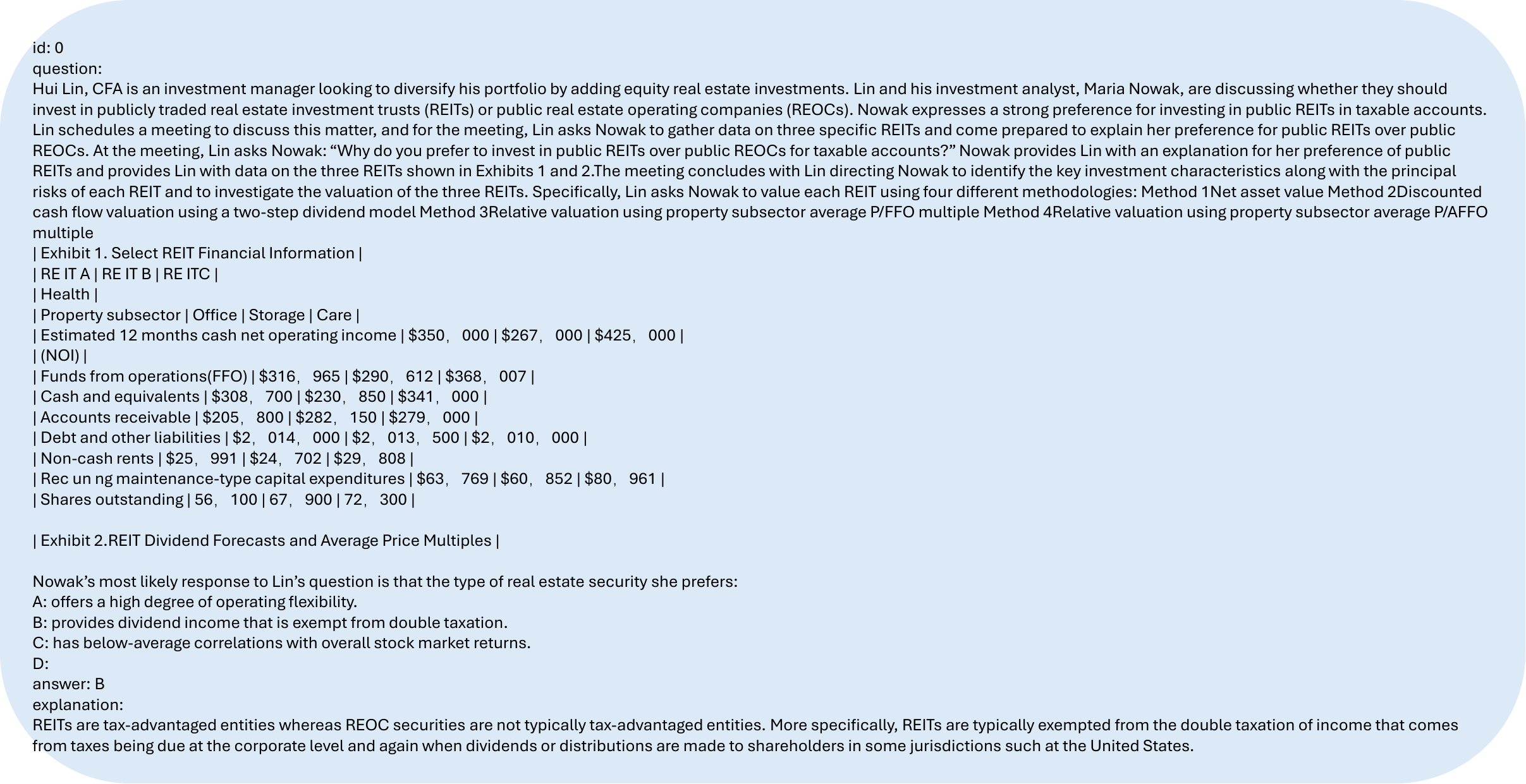}
    \caption{An example of CFA Level 2 problem.}
    \label{fig:cfa_l2}
\end{figure}

\paragraph{Chartered Financial Analyst} The CFA dataset comprises Level 1 and Level 2 examination data, covering a wide range of topics such as ethics and professional standards, quantitative methods, economics, financial reporting and analysis, corporate finance, equity investments, fixed income, derivatives, alternative investments, and portfolio management. This dataset thoroughly assesses LLMs' understanding of economics, finance, and asset management, as well as their ability to analyze real financial cases. Examining the levels of examination, the CFA Level 1 primarily consists of single-choice questions that generally do not involve complex charts, presenting relatively simple items that focus on assessing the models' grasp of fundamental financial knowledge. This forms the foundation for building advanced financial understanding. On the other hand, the CFA Level 2 features case study questions that typically provide a detailed case background along with related graphical data, and then pose several multiple-choice questions based on the case content. These questions are comparatively more complex, emphasizing the generic models' analytical, judgmental, and decision-making capabilities, especially in handling intricate scenarios and multi-variable situations. Example CFA problems are shown in Figure \ref{fig:cfa_l1} and \ref{fig:cfa_l2}.

\subsection{Data Sources}

In the preparation of our study materials for the CPA examination, we have sourced an extensive collection of exam questions from the authoritative Chinese CPA website \textit{ZhanLiuJiang}. This collection encompasses a wide array of simulated exam questions from recent years, in addition to actual exam questions released by official sources. Conversely, for the CFA examination, due to the absence of a centralized, authoritative data source for exam questions, we have resorted to aggregating questions from a diverse range of third-party individuals and institutions.

\subsection{Data Processing}

The original data set is stored in JSON format. The original examination papers for CFA Level 2 often include extensive tables or diagrams as supplementary information to the questions. In the JSON data, these tables are represented by URLs linking to images. To extract and convert these tables into a structured format, we utilized the table recognition API provided by Alibaba Cloud (https://api.aliyun.com/document/ocr/2019-12-30/RecognizeTable). This process involves a preliminary step to determine whether the input image can be recognized as a table. Upon successful recognition, the API returns the table information in JSON format. We then convert this JSON data into Markdown format, enabling us to seamlessly integrate the structured table information into our dataset by replacing the corresponding image URLs in the questions. An example of this conversion process is illustrated in Figure \ref{fig:cfa_l2}.

Drawing inspiration from the dataset format recommended by \cite{huang2024ceval}, we have meticulously segmented each dataset into development (dev), validation (val), and test sets, with the questions for each subject meticulously organized within separate CSV files. The development set serves as a foundational corpus for building prompts, complete with a few illustrative examples designed to support in-context learning. For each subject, we provide five questions, each accompanied by a question stem, four options (A, B, C, D), the correct answer, and a detailed explanation. The validation set, on the other hand, omits the explanation component, retaining only the correct answers. The test set further streamlines this by excluding both answers and explanations. In an effort to stimulate rapid development and iterative enhancements within the community, we have strategically placed the test set predominantly within the val folder, providing access to the ground truth for each question.

\subsection{Data Statistics}

Here, we conduct a statistical analysis of the different categories within IDEA-FinBench, namely, the number of questions per subject for CPA or CFA, as well as the total number of questions. For the sake of table conciseness, we do not distinguish between the different types of questions for CPA and CFA, which can refer to Table \ref{table:finkbench_statistics}.

\begin{table}[ht]
    \centering
    \begin{tabular}{llr}
    \toprule
    \textbf{Category} & \textbf{Subject} & \textbf{\#Questions} \\
    \midrule
    CPA & Accounting & 648 \\
    & Auditing & 641 \\
    & Economic Law & 319 \\
    & Financial Management & 407 \\
    & Strategy & 126 \\
    & Tax Law & 475 \\
    \textit{Total for CPA} & & \textit{2616} \\
    \midrule
    CFA & Alternative Investments & 103 \\
    & Corporate Finance & 155 \\
    & Derivatives & 118 \\
    & Economics & 222 \\
    & Equity & 225 \\
    & Ethical and Professional Standards & 81 \\
    & Financial Reporting and Analysis & 359 \\
    & Fixed Income & 253 \\
    & Portfolio Management & 229 \\
    & Quantitative Method & 256 \\
    \textit{Total for CFA} & & \textit{2001} \\
    \midrule
    \multicolumn{2}{l}{\textbf{Grand Total}} & \textbf{4617} \\
    \bottomrule
    \end{tabular}
    \caption{Number of Questions per Subject for CPA and CFA Categories}
    \label{table:finkbench_statistics}
\end{table}

\section{Experimental Settings}

In the following chapters, we will introduce the experimental settings of IDEA-FinBench, covering the evaluation setup, the model list, and the experimental results.

\subsection{Evaluation Setup}

The quality of generation by LLMs is influenced by various factors. Beyond the adjustments in decoding strategies, such as temperature settings, the prompt itself is a focal point for content generation. For instance, when the model is encouraged to engage in step-by-step reasoning, a performance improvement is often observed \cite{wei2022chain}. Furthermore, the insertion of appropriate examples into the prompt of an LLM can activate its in-context learning capabilities \cite{brown2020language}.

\paragraph{Few-shot or Zero-shot Learning} In the few-shot learning scenario, we utilize examples from each subject in the development set as references for the current question, tailoring both the instructions and responses to ensure that the model's output conforms to a specific style. In contrast, in the zero-shot scenario, the model must directly answer the question without any contextual references, making it more suitable for tracking performance improvements of LLMs that have undergone instruction-tuning compared to their base versions.

\paragraph{Chain-of-Thought or Answer-Only}
The chain-of-thought mode expects the LLM to deconstruct its reasoning steps when tackling a problem, especially in scenarios involving complex reasoning. Therefore, the model tends to provide a lengthy sequence of thought processes before delivering its final answer. On the other hand, the answer-only mode adopts a greedy decoding approach, where the LLM's next token to be generated is restricted to a fixed vocabulary space (e.g., options A, B, C, D in IDEA-FinBench) after receiving the input instruction. This approach directly captures the model's response tendency in solving the problem as the final answer.

\subsection{Models} 

To comprehensively assess the capabilities of various LLMs in mastering, understanding, and applying financial knowledge under our IDEA-FinBench, we conducted evaluations on up to 21 different language models. Among these, some have only made their access interfaces available, whereas the majority have disclosed the network weights of the models and provided public download permissions within the community. Additionally, the predominant share of these LLMs is geared towards general domains. However, a smaller segment, focused on vertical domains—specifically those LLMs that have undergone additional training with financial corpora—has also been included in our assessment. Detail information can refer to Table \ref{table:finkbench_models}.

\begin{table}[ht]
    \centering
    \begin{tabular}{lccc}
    \toprule
    \textbf{Model}              & \textbf{Size} & \textbf{Access} & \textbf{Base Model}       \\
    \midrule
    ChatGPT                     & -             & API             & -                         \\
    GPT-4                       & -             & API             & -                         \\
    LLaMA-2-chat                & 7B, 13B       & Weights         & LLaMA-2                   \\
    Chinese-Alpaca-2            & 7B, 13B       & Weights         & LLaMA-2                   \\
    ChatGLM3-Base               & 6B            & Weights         & -                         \\
    ChatGLM3-6B                 & 6B            & Weights         & ChatGLM3-6B-Base          \\
    Baichuan2                   & 7B, 13B       & Weights         & -                         \\
    Baichuan2-Chat              & 7B, 13B       & Weights         & Baichuan2                 \\
    Qwen                        & 7B, 14B       & Weights         & -                         \\
    Qwen-Chat                   & 7B, 14B       & Weights         & Qwen                      \\
    Yi                          & 6B            & Weights         & -                         \\
    Yi-Chat                     & 6B            & Weights         & Yi                        \\
    Tongyi-Finance              & 14B           & Weights         & Qwen-14B                  \\
    Tongyi-Finance-Chat         & 14B           & Weights         & Tongyi-Finance-14B        \\
    DISC-FinLLM                 & 13B           & Weights         & Baichuan2-13B-Chat        \\
    \bottomrule
    \end{tabular}
    \caption{Models evaluated in IDEA-FinBench.}
    \label{table:finkbench_models}
\end{table}

\paragraph{GPTs} OpenAI has developed ChatGPT \cite{OpenAI2022ChatGPT}, one of the world's most popular language model-based dialogue systems, as well as its advanced version, GPT-4 \cite{openai2023gpt4}. To this day, GPT-4 continues to lead on various leaderboards for natural language understanding and generation tasks, enjoying widespread popularity. The training of the model employed rigorous alignment techniques to learn human preferences \cite{ouyang2022training}. However, the model's weight files have not been made public, and there are only web and API available for being accessed.

\paragraph{LLaMA-2} MetaAI's contributions to the open-sourcing community for LLMs are significant, with the release of LLaMA and LLaMA 2 serving as milestones \cite{touvron2023llama, touvron2023llama2}. In comparison to the original Transformer's Decoder architecture \cite{vaswani2017attention}, LLaMA adopts RMSNorm for pre-normalization, SwiGLU as the activation function, and RoPE for positional embedding. LLaMA 2, on the other hand, replaces the original attention mechanism with Grouped-query Attention (GQA) and also expands the pre-training corpus. Our usage of LLaMA2-Chat is also aligned based on labeled data that conforms to human preferences.

\paragraph{Chinese-Alpaca-2} As an extension of the LLaMA 2 model in Chinese scenarios, Chinese-Alpaca expands the Chinese vocabulary based on the original LLaMA model weights and undergoes secondary pre-training and alignment based on Chinese instruction data \cite{cui2023efficient}. This enhances its semantic understanding and instruction-following abilities in Chinese contexts, making a pioneering contribution to the Chinese NLP community.

\paragraph{ChatGLM-3} Introduced by Zhipu AI and Tsinghua University, chatglm3-6B achieved the strongest performance among LLMs with less than 10 billion parameters in various evaluation rankings upon its release \cite{ChatGLM3}. It natively supports scenarios such as function calling, code interpretation, and agent tasks.

\paragraph{Baichuan-2} Developed by Baichuan Intelligent Technology, Baichuan 2 was pre-trained using 2.6 trillion tokens and is available in two versions: 7 billion and 13 billion parameters \cite{yang2023baichuan}. Building on the foundation of its predecessor, it expanded the vocabulary and context length, and improved the model's performance in multilingual contexts.

\paragraph{Qwen} Qwen, short for Tongyiqianwen, is open-sourced by Alibaba Cloud \cite{bai2023qwen}. It is pre-trained using over 3 trillion tokens of high-quality corpus, and demonstrates strong competitiveness among models of similar scale. Additionally, Qwen employs SFT and RLHF \cite{ouyang2022training} techniques for model alignment, resulting in the Chat series, which possess instruction-following and interactive capabilities.

\paragraph{Yi} As a model based on the native LLaMA architecture, Yi is dedicated to achieving leadership in bilingual language models \cite{young2024yi}. The Yi series of language models demonstrate strong capabilities in language cognition, common sense reasoning, and reading comprehension.

\paragraph{DISC-FinLLM} Based on the general-domain Baichuan-13B-Chat, DISC-FinLLM is meticulously fine-tuned using high-quality financial datasets \cite{chen2023disc}. Employing LoRA technology \cite{hu2021lora} and training on different sets of instructions, DISC-FinLLM is dedicated to building a multi-expert intelligent financial system, providing users with professional, intelligent, and comprehensive financial advisory services.

\paragraph{Tongyi-Finance} Tongyi-Finance \cite{tongyi_finance_14b_chat} uses Qwen as its base model and conducts continue pre-training using financial industry language data to enhance its knowledge and application capabilities in the financial domain. It covers abilities such as financial knowledge Q\&A, text classification, information extraction, text generation, reading comprehension, logical reasoning, multi-modality, and coding.

\paragraph{IDEA-FinLLM} IDEA-FinLLM uses Yi-34B-Chat \cite{young2024yi} as the base model, and is fine-tuned using large-scale, multi-dimensional financial knowledge instructions based on the FinKER \ref{chp_finker} developed by IDEA Research. This enhancement improves the model's capabilities in knowledge retention, numerical calculation, logical reasoning, and reading comprehension across financial scenarios.

\section{Result \& Analysis}

In Table \ref{table:finbench_leaderboard}, we present the comprehensive performance of 21 LLMs on IDEA-FinBench, evaluated by averaging the accuracy of each model across four categories: CPA with single answer (CPA-SA), CPA with multiple answers (CPA-MA), CFA Level 1 (CFA-L1), and CFA Level 2 (CFA-L2). In Table \ref{table:finbench_leaderboard}, "Random" serves as the baseline, representing the probability of randomly selecting an answer from the options available for each question. It is important to note that CPA questions have four options (A, B, C, D), resulting in a random accuracy rate of 25.00\% for CPA-SA and 10\% for CPA-MA (when randomly selecting a combination of answers). In contrast, CFA questions have three options (ABC), leading to a random accuracy rate of 33.33\%.

GPT-4 demonstrated remarkable leadership  across almost all categories and subjects, particularly in the English-medium CFA exams. LLaMA showed weaker performance in Chinese questions but ranked closer to the middle in English questions. Chinese-Alpaca improved in CPA questions due to its domain expansion in Chinese corpora, but this came at the expense of its capabilities in the English domain. Observations based on the evaluation results of ChatGLM, Baichuan, and Qwen reveal that, generally, base models exhibit a slight advantage over their corresponding chat models, but the loss incurred due to instruction-tuning is not significant. Furthermore, expanding the parameter size of the same model architecture leads to a stable increase in accuracy, enhancing problem-solving abilities. However, simply increasing the parameter size is insufficient to overcome performance variations caused by different model selections and pre-training corpus construction strategies among organizations. For instance, the Baichuan2-13B series lagged behind or matched the Qwen-7B series across various categories, while the Yi-6B series, with the smallest model size in IDEA-FinBench, demonstrated exceptional performance, even surpassing GPT-4 in CPA questions, which dominated the CFA categories.

\begin{table}[ht]
    \centering
    \begin{tabular}{lrrrr}
    \toprule
    \textbf{Model} & \textbf{CPA-SA} & \textbf{CPA-MA} & \textbf{CFA-L1} & \textbf{CFA-L2} \\
    \midrule
    Random                  & 25.00  & 10.00  & 33.33  & 33.33  \\
    \midrule
    ChatGPT             & 42.64  & 26.88  & 66.48  & 42.17  \\
    GPT-4                   & 62.38  & 45.27  & \textbf{84.26}  & \textbf{60.84}  \\
    IDEA-FinLLM              & \textbf{78.71} & \textbf{62.35} & 75.49 & 53.87 \\
    Llama-2-7b-chat         & 29.77  & 4.20   & 45.82  & 28.46  \\
    Llama-2-13b-chat        & 29.92  & 9.37   & 50.00  & 36.30  \\
    chinese-alpaca-2-7b     & 33.03  & 7.88   & 40.66  & 23.34  \\
    chinese-alpaca-2-13b    & 36.00  & 10.51  & 46.64  & 31.63  \\
    chatglm3-6b-base        & 49.79  & 14.89  & 58.28  & 37.65  \\
    chatglm3-6b             & 41.80  & 20.84  & 42.62  & 32.98  \\
    Baichuan2-7B-Base       & 42.50  & 9.72   & 50.90  & 29.37  \\
    Baichuan2-7B-Chat       & 41.80  & 13.57  & 42.95  & 31.17  \\
    Baichuan2-13B-Base      & 45.90  & 20.32  & 56.56  & 42.77  \\
    Baichuan2-13B-Chat      & 45.40  & 14.45  & 51.31  & 39.31  \\
    DISC-FinLLM             & 38.68  & 9.98   & 43.77  & 30.12  \\
    Qwen-7B                 & 49.65  & 19.96  & 56.56  & 39.46  \\
    Qwen-7B-Chat            & 47.17  & 24.78  & 52.70  & 40.81  \\
    Qwen-14B                & 59.48  & 18.04  & 63.61  & 47.44  \\
    Qwen-14B-Chat           & 58.20  & 36.43  & 59.26  & 46.99  \\
    Tongyi-Finance-14B      & 51.34  & 28.37  & 63.44  & 45.78  \\
    Tongyi-Finance-14B-Chat & 49.50  & 15.50  & 58.28  & 41.72  \\
    Yi-6B                   & 64.43  & 40.63  & 60.49  & 26.20  \\
    Yi-6B-Chat              & 63.22  & 47.20  & 53.36  & 28.46  \\
    \bottomrule
    \end{tabular}
    \caption{Average accuracy (\%) on the test set. The "SA" in "CPA-SA" column refers to CPA questions with a single answer, the "MA" in "CPA-MA" column refers to CPA questions with multiple answers. Additionally, the "L1" in "CFA-L1" column refers to questions from CFA Level 1, and the "L2" in "CFA-L2" column refers to questions from CFA Level 2.}
    \label{table:finbench_leaderboard}
\end{table}

Finally, we discuss observations on Financial LLMs that underwent secondary training on financial corpora based on base models. Surprisingly, these vertical-specific models did not achieve the anticipated improvements on IDEA-FinBench, despite being more domain-adapted compared to general-purpose models. Further experiments are needed for a deeper investigation.

\section{Conclusion}

In this chapter, we present IDEA-FinBench, an innovative benchmark designed to assess financial knowledge in LLMs by utilizing questions from two globally recognized and authoritative financial professional exams. The benchmark encompasses questions in both Chinese and English, four types of question formats, and spans sixteen financial disciplines, thereby providing a comprehensive evaluation of LLMs' ability to directly address exam questions pertinent to the finance sector. Moreover, we introduce a modular evaluation suite that can integrate external datasets, allowing for flexible customization of evaluation modes and interfaces with various LLMs. This feature enhances the adaptability and scalability of the evaluation framework, making it a versatile tool for assessing financial knowledge in LLMs.

In this study, our experimental results demonstrate that GPT-4 exhibited exceptional performance across nearly most of categories and subjects, particularly in the English-medium CFA exams. According to our observations, generally, base models have a slight advantage over their corresponding chat models, with the loss incurred due to instruction-tuning being insignificant. Additionally, increasing the parameter size of the same model architecture leads to a stable increase in accuracy, enhancing problem-solving abilities. However, merely increasing the parameter size is insufficient to overcome performance variations caused by different model selections and pre-training corpus construction strategies among organizations. Finally, observations on Financial LLMs that underwent secondary training on financial corpora based on base models revealed that these vertical-specific models did not achieve the anticipated improvements on IDEA-FinBench, despite being more domain-adapted compared to general-purpose models.
\newpage

\chapter{IDEA-FinKER: Financial Knowledge Enhancement Framework}
\label{chp_finker}
Based on observations in Chapter \ref{chp_finkbench}, it is evident that large language models (LLMs) still face significant challenges in domain adaptation within the financial area. For instance, further pre-training and fine-tuning of foundational language models using financial corpora and instruction datasets did not yield the expected improvements on IDEAFinBench, and even resulted in a noticeable decline in performance. Upon further investigation, it was discovered that most publicly available financial LLMs primarily focus on instruction-following capabilities, such as natural language understanding tasks within the financial domain, covering entity recognition, summarization, event extraction, and more \cite{chen2023disc, li2023cfgpt, tongyi_finance_14b_chat}. It is realized that the paradigm of financial knowledge injection warrants further exploration, whether from the perspective of in-context learning or supervised fine-tuning.

We introduce IDEA-FinKER, a \textbf{\underline{F}}inancial \textbf{\underline{K}}nowledge \textbf{\underline{E}}nhancement f\textbf{\underline{R}}amework, designed to facilitate the rapid adaptation of general LLMs to the financial domain without incurring the high costs associated with external pre-training. This framework is supported by a meticulously cleaned and constructed comprehensive database of Chinese financial exam questions, which incorporates support embedding similarity retrieval. IDEA-FinKER underpins the development of a retrieval-based few-shot learning method for real-time context-level knowledge injection, termed soft-injecting paradigm of knowledge. Additionally, we have developed a high-quality set of financial knowledge instructions for fine-tuning any general LLM, referred to as hard-injecting paradigm of knowledge. Empirical evidence demonstrates that IDEA-FinKER significantly enhances the expert capabilities of LLMs within the financial domain, notably improving their performance on the IDEAFinBench, especially in the segment pertaining to Chinese exam questions like CPA.

\section{Introduction}

This observation can be made by examining the training processes of some Chinese community open-access LLMs specific to the financial domain. The dataset used for fine-tuning DISC-FinLLM \cite{chen2023disc} includes a large number of instruction-following tasks, encouraging the model to enhance its instruction-following ability for language understanding tasks such as sentiment analysis, intent recognition, entity extraction, and retrieval-based question answering. However, the lack of financial knowledge injection results in a decline in performance on IDEAFinBench and FinEval after full-parameter fine-tuning. Similarly, the supervised fine-tuning dataset for CFGPT \cite{li2023cfgpt} also heavily favors tasks like text summarization, sentiment classification, and entity recognition. In contrast, Tongyi-Finance \cite{tongyi_finance_14b_chat}, despite using high-quality financial corpora for incremental pre-training and instruction fine-tuning, potentially causes significant damage to the original parameter structure determined by pre-training. These projects might be overly focused on training LLMs to be text assistants that follow a fixed work paradigm, rather than learning and mastering financial knowledge as experts.

Our contributions can be summarized as follows:

\begin{enumerate}

    \item \textbf{A Novel Framework for Financial Domain Adaptation:} IDEA-FinKER represents a unique approach to adapting general LLMs to the financial domain. Unlike traditional methods that rely heavily on extensive pre-training with financial corpus, IDEA-FinKER facilitates rapid adaptation without incurring high costs, making it a cost-effective solution for enhancing the financial expertise of LLMs.
    
    \item \textbf{Investigation on Knowledge Injection Paradigms:} We explore two distinct paradigms for knowledge injection in LLMs: the soft-injecting paradigm, which employs retrieval-based few-shot learning for real-time context-level knowledge injection, and the hard-injecting paradigm, which involves fine-tuning LLMs with a set of high-quality financial knowledge instructions.
    
    \item \textbf{Analysis on IDEA-FinKER's Performance on Knowledge Injection:} Our empirical evaluation demonstrates that IDEA-FinKER significantly improves the performance of LLMs on financial tasks. It's observed that IDEA-FinKER achieves the best performance when integrate both the two paradigms together on different base models.

\end{enumerate}

\section{Methodology}

In this chapter, we will introduce the methodology of IDEA-FinKER. First, we collected and organized FinCorpus from the internet. Our framework includes two paradigms of knowledge injection: the soft-injecting paradigm and the hard-injecting paradigm. To adapt to different paradigms, our FinCorpus needs to be processed accordingly and constructed into different formats for context insertion or fine-tuning.

\subsection{FinCorpus}
\begin{figure}[htbp]
    \centering
\includegraphics[width=1.0\linewidth]{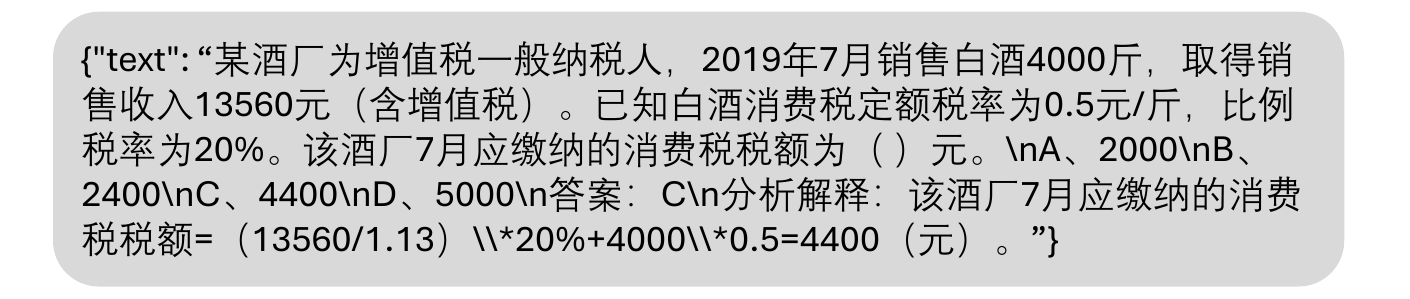}
    \caption{An example of problem in financial corpus.}
    \label{fig:fin_corpus}
\end{figure}

We collect FinCorpus, which is a set of financial questions exceeding 400M in size from the internet, consisting of approximately 500,000 questions in Chinese, including multiple options, covering finance, economics, insurance, certifications, etc., stored in JSONL format, with an example provided in Figure \ref{fig:fin_corpus}. The statistics of FinCorpus is provided in Table \ref{table:stat_fincorpus}.

\begin{table}[ht]
\centering
    \begin{tabular}{ll}
    \toprule
    \textbf{Metric}                 & \textbf{\#Counts}   \\
    \midrule
    Total items                     & 498,043          \\
    Unique items                    & 336,897          \\
    Average text length             & 255.41           \\
    Minimum text length             & 69               \\
    Maximum text length             & 11,833           \\
    Average question length         & 115.74           \\
    Average answer length           & 130.32           \\
    \midrule
    \multicolumn{2}{l}{\textbf{Problems with different options}}   \\
    \midrule
    With 2 options                       & 24,604           \\
    With 3 options                       & 1,214            \\
    With 4 options                       & 258,037          \\
    With 5 options                       & 47,427           \\
    Others                          & 5,615            \\
    \bottomrule
    \end{tabular}
\caption{Statistics of FinCorpus.}
\label{table:stat_fincorpus}
\end{table}

The data cleaning process are performed as follow. At first, we split each item into its question and answer parts by using regular expressions to match patterns similar to "Answer: " as the delimiter for splitting, to obtain the question and answer. Next, we perform de-duplication to minimize the data size as much as possible and retain only the non-repetitive parts. We choose to remove irrelevant characters from the question field of each item, retain only the Chinese characters, and truncate the first 30 characters, using them as unique identifiers. As a result, repetitive items will be filtered out. Additionally, we use rules to remove irrelevant prefixes and suffixes. Finally, we obtain approximately 300,000 financial questions, each containing multiple options, the correct answer, and relevant explanations.

\subsection{Soft-Injecting Paradigm}

According to the observations from \cite{brown2020language}, as the scale of parameters and corpus for pre-trained models expands, LLMs begin to exhibit the capability of in-context learning. This phenomenon allows LLMs to adapt to new tasks or understand novel instructions based solely on the context provided within the input text, such as a few examples given in the form of demonstration, without the need for explicit retraining or fine-tuning. Therefore, directly injecting financial knowledge into the context of LLMs based on retrievers can be regarded as an effective soft-injecting paradigm.

Following the definition from \cite{dong2022incontext}, LLMs calculate the likelihood of the next token to concatenate a potential answer, conditioned on the provided context with a problem included. Consider a problem \(P\) for which we have a set of candidate answers \(O = \{a_1, \ldots, a_n\}\), each candidate answer is associated with corresponding textual data. A pre-trained language model \(M\) to identify the answer \(A\) with the highest confidence score conditioned on a set of demonstrations \(D\). The set \(D\) comprises a system instruction \(I\) for executing the task and \(k\) demonstration examples. Each example \(e_i\) is a triplet consisting of a problem, options, and an answer, i.e., \(e_i = (p_i, o_i, a_i)\). Hence, the demonstration set is formulated as 
\begin{equation}
    D = \{I, e_1, \ldots, e_k\} = \{I, (p_1, o_1, a_1), \ldots, (p_k, o_k, a_k)\}, 
\end{equation}
where \(k\) may vary, corresponding to zero-shot, one-shot, and few-shot scenarios.

In scenarios where the problem involves choosing one option from a set of options such as "A, B, C, D", the model predicts the token with the highest confidence score. To quantify the likelihood of a candidate answer \(a_i\), we employ a scoring function \(f\) that takes the entire sequence as input to the model \(M\), yielding:
\begin{equation}
    P(a_i \mid P) = f(a_i, P, D, M),
\end{equation}
where \(A \in O = \{a_1, \ldots, a_n\}\). The final predicted answer \(\hat{A}\) is then determined by selecting the candidate with the highest confidence score:
\begin{equation}
    \hat{A} = \arg\max_{a_i \in O} f(a_i, P, D, M).
\end{equation}

The scoring function \(f\) leverages the examples in \(D\) to learn the mapping between inputs and labels, thereby assessing the likelihood of correctness for each candidate answer. In cases where candidate answers are combinations of two or more options, say \(b_i\) is a subset of \(O\) with \(b_i = \{a_i, \ldots, a_m\}\), the set of all possible combinations is denoted as \(A \in \{b_1, \ldots, b_z\}\), where \(z = 2^n - n - 1\). The final predicted answer in such scenarios is obtained by maximizing the joint probability of the option combination:
\begin{align}
    \hat{A} &= \arg\max_{b_i \in O} f(b_i, P, D, M) \nonumber \\
    &= \arg\max_{\{a_i, \ldots, a_m\} \in O} f(\{a_i, \ldots, a_m\}, P, D, M) \nonumber \\
    &= \arg\max_{\{a_i, \ldots, a_m\} \in O} \prod_{j=i}^{m} f(a_j, P, D, M).
\end{align}

The definition from \cite{xie2021bayesian} states that in-context learning is essentially the pre-trained language model performing implicit Bayesian inference, that is, inferring the shared prompt concept among the given examples to complete the current task. Therefore, the improvement in inference performance is inevitably related to the quality of the examples provided in the context. When the examples used as demonstrations are fixed, we hope that these samples have a wide range of representativeness and are as relevant as possible to each input question. To achieve this, human experts in the relevant field need to intervene and manually write high-quality samples as references for insertion into the context \cite{singhal2023clinical}. However, this not only requires a certain amount of human resources, but it is also difficult to judge whether these small samples accurately provide valuable information for any sample input into the LLM. The empirical study from \cite{liu2021makes, nori2023generalist} also demonstrates that, although they are consistent in format, cases that are more similar to the current problem, which is usually measured by text similarity, serve as demonstrations and are inserted into the context of the LLM. Compared to samples that are far away, they can bring about a more significant improvement in accuracy.

\begin{algorithm}
\label{alg:rbfl}
\caption{Retrieval-based Few-shot Learning}
    \begin{algorithmic}[1]
    \Require The problem $p$, Knowledge base $\phi$, Number of shot $K$
    \Ensure The answer $\alpha$
    \State $\texttt{Encoder} \gets \text{function to encode text into embedding}$
    \State $\texttt{Index} \gets \text{build\_index}(\phi, \texttt{Encoder})$
    \State $\texttt{ctx} \gets \{\}$
    \State $t \gets 0$
    \State \textbf{repeat}
        \State \hspace{\algorithmicindent}$E \gets \text{retrieve}(\phi, \texttt{Index}, \texttt{Encoder}, p)$
        \State \hspace{\algorithmicindent}$\text{shot}  \gets \text{build\_shot}(E)$
        \State \hspace{\algorithmicindent}$\texttt{ctx}  \gets \texttt{ctx} \cup \text{shot}$
        \State \hspace{\algorithmicindent}$\phi \gets \phi - E$
        \State \hspace{\algorithmicindent}$t \gets t + 1$
    \State \textbf{until} {$t \geq K$}
    \State $\texttt{ctx} \gets \texttt{ctx} \cup p$
    \State $\texttt{prompt} \gets \text{build\_prompt}(\text{ctx})$
    \State $\alpha \gets \texttt{LLM}\text{.generate}(\texttt{prompt})$
    \State \Return $\alpha$
    \end{algorithmic}
\end{algorithm}

\begin{figure}[htbp]
    \centering
\includegraphics[width=1.0\linewidth]{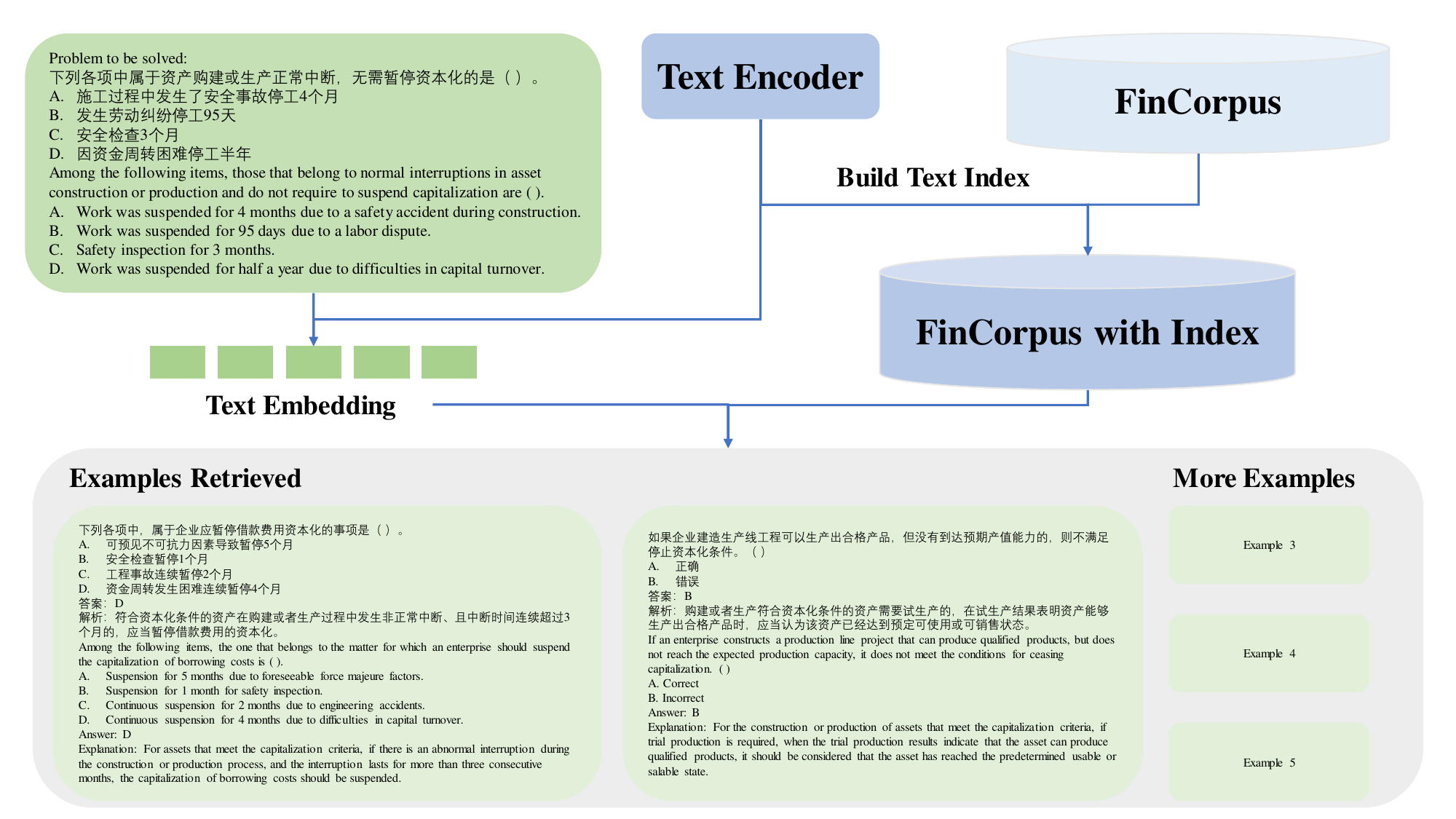}
    \caption{An example of the working pipeline of Retrieval-based Few-shot Learning.}
    \label{fig:rbfl}
\end{figure}

The Retrieval-based Few-shot Learning (RBFL) algorithm employs a pre-built external knowledge base, FinCorpus, to augment the learning process. Initially, an encoder function is established to convert text into embeddings, and an index is constructed from FinCorpus using this encoder for efficient information retrieval. For a specified number of shots $K$, the algorithm retrieves relevant embeddings from FinCorpus based on the problem $p$, constructs a "shot" from these embeddings, and accumulates this shot in the context $\texttt{ctx}$, while ensuring that retrieved embeddings are removed from the knowledge base to prevent redundancy. After completing $K$ iterations of retrieval, the problem statement $p$ is added to the context, and a prompt is constructed from this enriched context. The prompt is then fed into a LLM, which generates the answer $\alpha$ based on the provided information. The algorithm ultimately returns this answer as the solution to the problem, leveraging the few-shot learning approach with the support of the external knowledge base FinCorpus. The formal algorithmic process is defined in Alg. \ref{alg:rbfl}, and a visual chart of the pipeline is given in Figure \ref{fig:rbfl}.

\subsection{Hard-Injecting Paradigm}

We initially propose a categorization criterion for financial instructions aimed at ensuring that the data covers all critical aspects of the financial sector. This standard allows the model to more accurately comprehend and process complex financial information. These instructions not only encompass a wide range of financial knowledge but also specifically consider the rigor of the instruction format and logic. Our goal is to guide the model to adapt to diverse financial knowledge application scenarios through these varied instructions. Additionally, we consciously adjust the output format to ensure that the model's responses are comprehensive, intelligent, and better aligned with users' preferences and needs.

Our financial instruction categorization criterion includes the following four categories:

\begin{enumerate}
    \item \textbf{Financial Knowledge Inquiry Instructions}: These instructions typically involve the examination of financial terms, concepts, entities, and nouns, requiring the model to have a strong memory capability to respond directly to user queries in a recitative manner. This type of instruction does not need complex reasoning steps, but the model should be able to cite the original text of relevant terms to provide more persuasive explanations.
    
    \begin{figure}[htbp]
        \centering
    \includegraphics[width=1.0\linewidth]{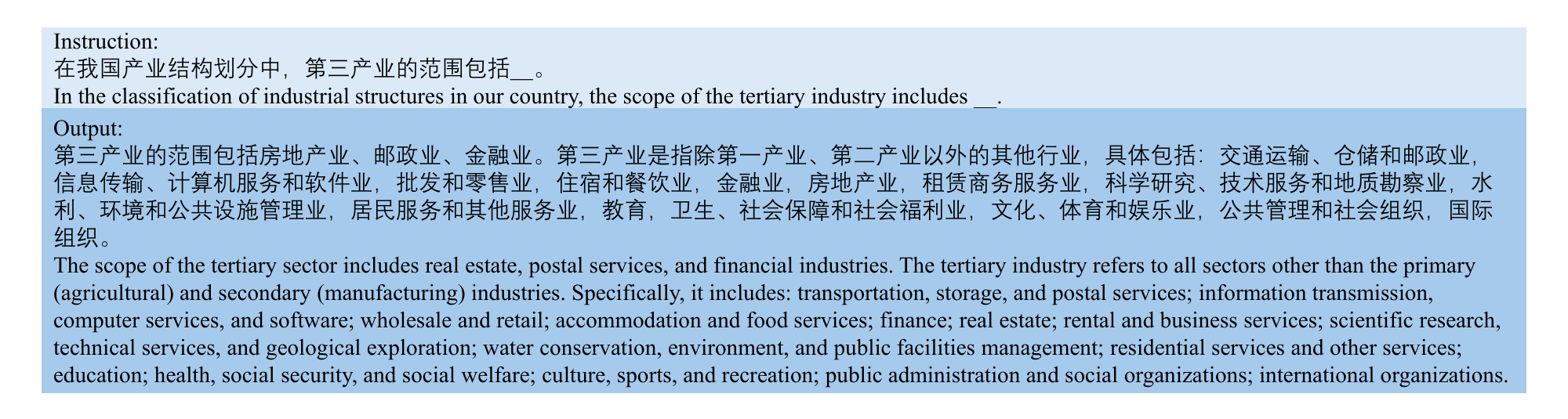}
        \caption{An example of the Financial Knowledge Inquiry Instruction.}
        \label{fig:knowledge_inquiry}
    \end{figure}

    \item \textbf{Financial Calculation and Reasoning Instructions}: These instructions require model to combine mathematical logic and reasoning capabilities with financial analysis skills. By adopting an inductive reasoning paradigm and integrating basic financial numerical concepts such as tax rates, growth rates, and interest rates, the model needs to establish rigorous calculation formulas and ultimately derive answers. In this scenario, the correctness and logic of the reasoning process take precedence over the numerical answer itself.

    \begin{figure}[htbp]
        \centering
    \includegraphics[width=1.0\linewidth]{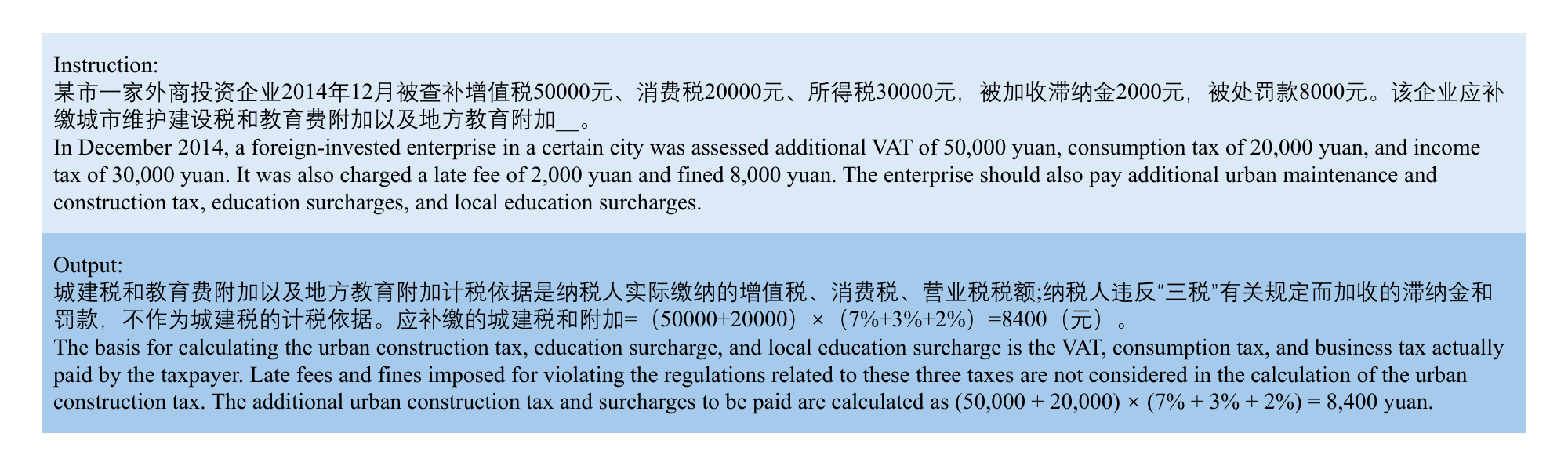}
        \caption{An example of the Financial Calculation and Reasoning Instruction.}
        \label{fig:financial_calculation}
    \end{figure}
    
    \item \textbf{Financial Reading Comprehension Instructions}: These instructions emphasize the need for the model to read and understand specific questions and each option to judge the logical relationships and make choices. This not only tests the understanding and memory of basic financial concepts but also requires the model to analyze real-life financial cases within the context of financial language, demonstrating more comprehensive and integrated financial logical thinking.

    \begin{figure}[htbp]
        \centering
    \includegraphics[width=1.0\linewidth]{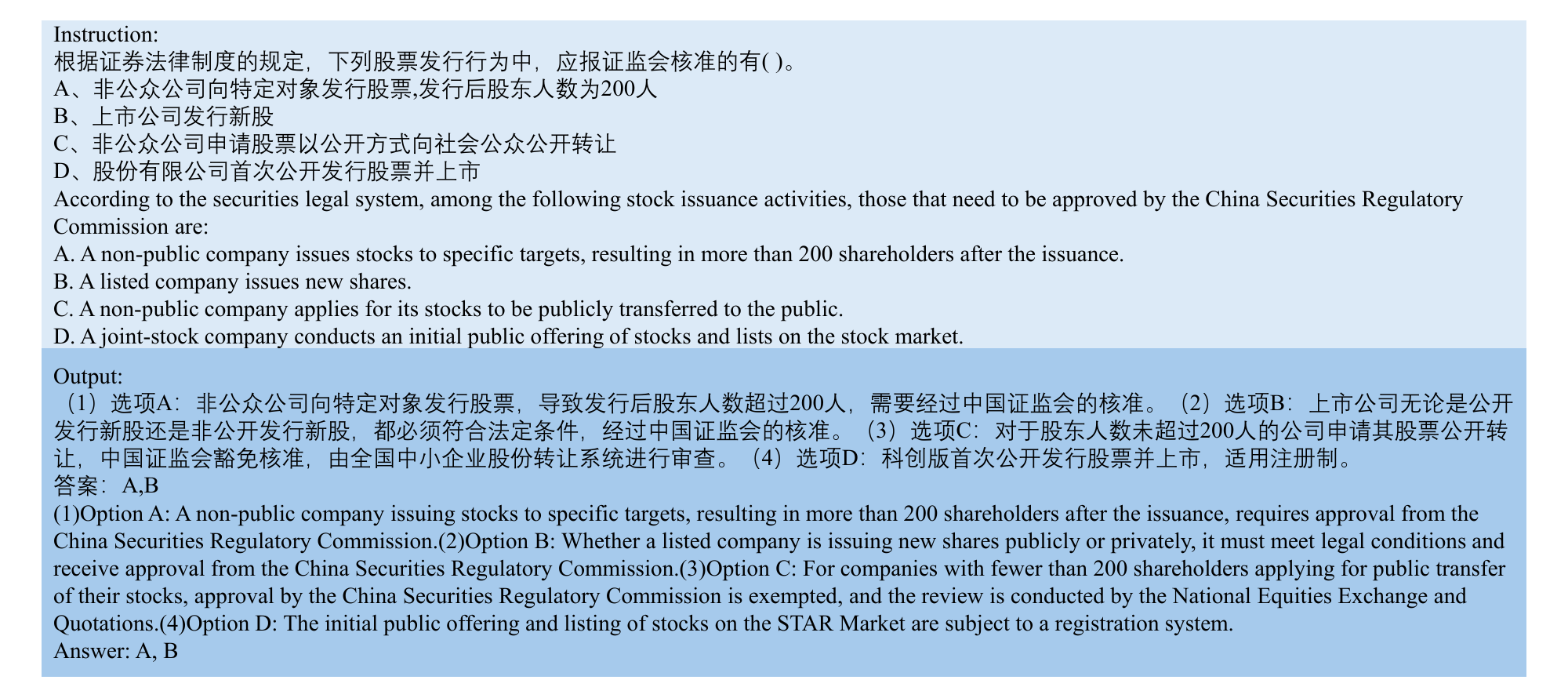}
        \caption{An example of the Financial Reading Comprehension Instruction.}
        \label{fig:financial_reading}
    \end{figure}
    
    \item \textbf{Financial Logical Judgment Instructions}: These instructions rigorously assess the arguments presented in the questions, evaluating their logical validity and applicability in financial scenarios. Compared to financial knowledge inquiry instructions that require the model to output corresponding knowledge points, this type encourages the model to develop critical thinking, imposing stricter constraints on potential model illusions and more keenly capturing possible logical flaws in the arguments.

    \begin{figure}[htbp]
        \centering
    \includegraphics[width=1.0\linewidth]{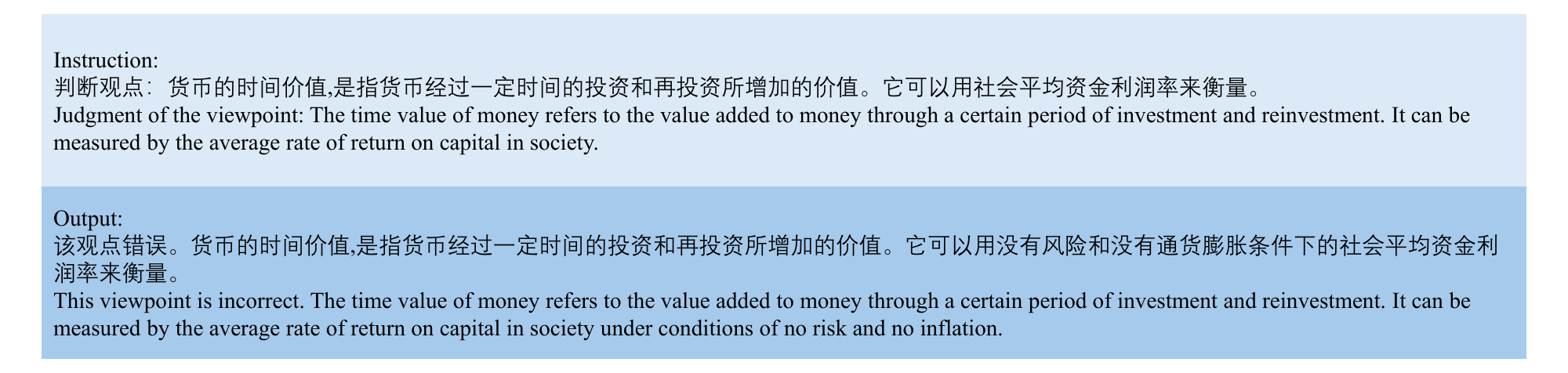}
        \caption{An example of the Financial Logical Judgment Instruction.}
        \label{fig:financial_logical}
    \end{figure}

\end{enumerate}

\section{Experimental Settings}

The first approach is the soft-injecting paradigm for financial knowledge infusion. Due to the large scale of the financial question bank used for dynamic context knowledge injection, we selected forty thousand questions as the candidate set. We utilize ChromaDB \cite{chroma_2023} as the vector database and employ the popular Chinese text embedding method bge-large-zh-v1.5 from BAAI \cite{baai_bge_large_zh_v1.5_2023} to calculate and store the text embeddings of all the questions in the candidate set. Next, we traverse the test set and use cosine similarity as the metric to retrieve the five most similar candidate questions as example questions for knowledge injection, thereby enhancing the performance of few-shot learning. The second approach is the Hard-Injecting Paradigm for financial knowledge injection, for which we need to use the LLaMA-Factory framework.

LLaMA-Factory \cite{Zheng_LlamaFactory_Unified_Efficient_2024} is a unified framework integrating various cutting-edge techniques for efficiently fine-tuning LLMs. It supports the fine-tuning of popular open-source LLMs such as LLaMA \cite{touvron2023llama}, Mistral \cite{jiang2023mistral}, Qwen \cite{bai2023qwen}, Yi \cite{young2024yi}, among others. Utilizing a scalable modular design, LLaMA-Factory facilitates additional training and alignment of base models, including incremental pre-training, instruction-supervised fine-tuning, reward model training, and RLHF. Moreover, LLaMA-Factory allows for the selective configuration of specific fine-tuning methods, such as LoRA \cite{hu2021lora} and QLoRA \cite{dettmers2023qlora}, and supports training acceleration algorithms like Flash-Attention \cite{dao2022flashattention}. As part of the LLM training suite, real-time monitoring and model deployment are also included.

In our experimental setup for hard-injecting paradigm, the fine-tuning process is executed on four GPUs. We use LoRA as the fine-tuning method, targeting some specific layers with a rank of 64 and an alpha value of 128. The training batch size per device is set to 4, with an equivalent size for evaluation batches. Gradient accumulation is applied every four steps to effectively increase the batch size. The pre-processing is handled by eight workers, and a cosine learning rate scheduler is used. Model checkpoints are saved and evaluated every 1000 steps, with the validation set comprising 0.01 of the training data. The evaluation strategy is based on fixed steps, and the best model is loaded at the end of training. The initial learning rate is set to 5e-5. The training is carried out for three epochs, with the option to plot the loss curve. The model is trained with bfloat16 precision for enhanced efficiency, and flash-attention is enabled for accelerated training. Finally, we will integrate both the soft and hard injecting paradigms to enhance the same model.

\paragraph{Baseline} For our baseline, we select the Baichuan2-7B-Chat \cite{yang2023baichuan}, Qwen-7B-Chat \cite{bai2023qwen} and Yi-6B-Chat \cite{young2024yi} models. We compare their vanilla models with their respective versions that have undergone the financial knowledge soft-injecting paradigm, the hard-injecting paradigm, and a combination of both injection methods.

\section{Result \& Analysis}

\begin{table}[ht]
    \centering
    \begin{tabular}{lll}
    \toprule
    \textbf{Models} & \textbf{CPA-SA} & \textbf{CPA-MA} \\
    \midrule
    \textit{Random} & 25.00 & 10.00 \\
    \midrule
    \textit{Baichuan2-7B-Chat} & 41.44 & 12.35 \\
    IDEA-FinKER & \textbf{49.79 (\textcolor{green}{+8.35})} & 20.05 (\textcolor{green}{+7.70}) \\
    \quad\quad w/o hard & 47.67 (\textcolor{green}{+6.23}) & 5.17 (\textcolor{red}{-7.18}) \\
    \quad\quad w/o soft & 44.48 (\textcolor{green}{+3.04}) & \textbf{28.46 (\textcolor{green}{+16.11})} \\
    \midrule
    \textit{Qwen-7B-Chat} & 47.10 & 24.26 \\
    IDEA-FinKER & \textbf{56.15 (\textcolor{green}{+9.05})} & \textbf{34.59 (\textcolor{green}{+10.33})} \\
    \quad\quad w/o hard & 53.54 (\textcolor{green}{+6.44}) & 32.31 (\textcolor{green}{+8.05}) \\
    \quad\quad w/o soft & 50.42 (\textcolor{green}{+3.32}) & 28.72 (\textcolor{green}{+4.46}) \\
    \midrule
    \textit{Yi-6B-Chat} & 63.01 & 47.20 \\
    IDEA-FinKER & \textbf{70.30 (\textcolor{green}{+7.29})} & \textbf{53.33 (\textcolor{green}{+6.13})} \\
    \quad\quad w/o hard & 67.68 (\textcolor{green}{+4.67}) & 48.77 (\textcolor{green}{+1.57}) \\
    \quad\quad w/o soft & 65.91 (\textcolor{green}{+2.90}) & 51.66 (\textcolor{green}{+4.46}) \\
    \bottomrule
    \end{tabular}
    \caption{Average accuracy (\%) on the test set, which is the CPA part of IDEAFinBench. The "SA" in "CPA-SA" column refers to CPA questions with a single answer, the "MA" in "CPA-MA" column refers to CPA questions with multiple answers. The percentages indicate the increase by green color or decrease by red color compared to the vanilla model.}
    \label{table:finker_result}
\end{table}

From Table \ref{table:finker_result}, it can be observed that in the majority of cases, models processed through IDEA-FinKER exhibit improvements compared to their vanilla counterparts. Here, the term 'vanilla model' refers to native models that directly employ zero-shot learning to complete tasks.

\paragraph{Both two paradigms show stable improvements.} We can observe that on Baichuan2-7B-Chat, Qwen-7B-Chat, and Yi-6B-Chat, both soft injecting and hard injecting, as well as the combined use of both methods, have achieved stable improvements in the subjects CPA-SA and CPA-MA compared to the vanilla models, with the only exception being a decline in Baichuan2-7B-Chat in the subject CPA-MA after soft injecting. The decrease might be attributed to Baichuan2-7B-Chat's relatively weaker few-shot learning capabilities compared to the base model, leading to a situation where the presence of multiple examples in the context adversely affected Baichuan2-7B-Chat's ability to follow instructions. In the six comparison cases provided for soft and hard injecting, hard injecting showed limited benefits, particularly for Yi-6B-Chat and Qwen-7B-Chat. The most significant enhancement with hard injecting was observed in the CPA-MA subject, reaching as high as 130.45\%. This improvement is linked to the vanilla Baichuan2-7B-Chat's inherent limitations in addressing multiple candidate answers. However, the hard-injecting paradigm demonstrated relatively limited benefits in other scenarios, especially regarding the enhancements for Yi-6B-Chat and Qwen-7B-Chat.

\paragraph{Combination of both paradigms achieves the best improvement.} When integrating both the soft-injecting and hard-injecting paradigms, models that underwent fine-tuning (hard injecting) combined with a retrieval-based few-shot learning method (soft injecting) generally led the pack in every test set.

\paragraph{IDEA-FinKER performs better on those models with lower baseline.} It was noted that the impact of IDEA-FinKER was more pronounced on models with weaker capabilities, such as Baichuan2-7B-Chat and Qwen-7B-Chat. Despite Yi-6B-Chat's commanding lead in the leaderboard, the enhancements from IDEA-FinKER were stable but not as significant.

\paragraph{The native performance gap is still hard to be bridged.} Observations from IDEAFinBench indicate that multiple factors, such as the architectural choices, parameter scales, and pre-training corpora of the base models, can lead to significant performance disparities on financial knowledge-related questions. Although IDEA-FinKER brings notable improvements to the models, it still struggles to compensate for the inherent performance disadvantages of the base models. For example, Baichuan2-7B-Chat, which adopts two paradigms for knowledge injection, can only approach the vanilla version of Qwen-7B-Chat. Similarly, the knowledge-injected Qwen-7B-Chat also fails to rival the native Yi-6B-Chat.

\section{Conclusion}

In this chapter, we introduce the IDEA-FinKER framework, a novel approach designed to enhance the financial knowledge of LLMs. Distinct from traditional methods that primarily depend on extensive pre-training with financial corpora, IDEA-FinKER facilitates rapid adaptation without incurring substantial costs, thereby presenting a cost-effective solution for augmenting the financial expertise of LLMs. This framework is particularly notable for its exploration of two distinct knowledge injection paradigms: the soft-injecting paradigm and the hard-injecting paradigm. The soft-injecting paradigm utilizes retrieval-based few-shot learning for real-time, context-level knowledge injection, while the hard-injecting paradigm involves fine-tuning LLMs with a curated set of high-quality financial instructions. Additionally, this chapter provides an analysis of IDEA-FinKER’s effectiveness in knowledge injection. Empirical evaluations demonstrate that IDEA-FinKER significantly enhances the performance of LLMs on financial tasks, particularly when integrating both injection paradigms across various base models.
\newpage

\chapter{IDEA-FinQA: Financial Question \& Answering System}
\label{chp_finqa}
Experimental results from previous chapters significantly indicate that pre-trained large language models (LLMs) emerge with profound knowledge reasoning capabilities due to their vast parameter scale, even from the specific perspective of the financial sector. Their performance in answering finance-related knowledge questions reflects the characteristics that LLMs have learned and generalized from financial textbook corpora during pre-training. However, LLMs are inherently limited by the scope of their training data, which fundamentally consists of a snapshot of internet corpora, including temporal and spatial dimensions. The spatial dimension can be determined by trainers during data consolidation, but the temporal dimension's limitations pose challenges for the application of LLMs requiring information specific to a date beyond the cutoff point. The method of updating model parameters also faces challenges, constrained by the fragile model parameter structure and the high costs associated with secondary pre-training and supervised fine-tuning (SFT). This leaves a considerable room for improvement in LLMs' ability to handle fact-based knowledge queries.

We first introduce FinFact, the first Chinese financial domain factual knowledge verification dataset. We have collected high-quality financial news from authoritative Chinese news websites, covering diverse themes such as macroeconomic policies, agricultural economics, real estate, China's A-shares, and industrialization, ensuring a rich content variety. We constructed question-answer pairs from structural and conversational perspectives, considering the factuality in dialogues. Subsequently, we present IDEA-FinQA, a financial question-answering system driven by LLMs. IDEA-FinQA adheres to a scheme of real-time knowledge injection and factual enhancement using external knowledge for LLMs. The system comprises three main modules: the data collector is responsible for collecting and integrating financial domain data, including data storage solutions, online and offline collection; the data querying module offers data search methods based on two types of search engines, traditional text-based indexing and popular embedding-based indexing, for multiple stages of recall and ranking; the driving force of IDEA-FinQA is four LLM-based agents, performing corresponding tasks given different prompts and contexts, including a query rewriter, intention detector, extractor and refiner, and a response generator. Our experiments demonstrate that IDEA-FinQA surpasses the majority of models in factual question-answering, even when the facts are from different years.

\section{Introduction}

As observed in Chapter \ref{chp_finker}, effectively enhancing the cognitive abilities of LLMs in the domain of finance—specifically their recognition, understanding, and mastery of authoritative financial knowledge—can be achieved through the injection of knowledge across multiple paradigms. This enhancement not only improves performance on financial examinations but also helps overcome illusions and heightens the model's factual awareness. However, factual knowledge in financial scenarios is not limited to key points from books, encyclopedias, and textbooks, such as those concerning "leveraged buyouts" or "return on investment." It also extends to temporally characterized facts in the real world \cite{hu2023large}.

There is no shortage of work modifying model parameters to inject dense factual knowledge. For instance, some studies, such as \cite{de2021editing, meng2022locating, fastedit}, target specific knowledge triplets by detecting activation sites of the key vector, encoding the fact relations in the value vector, and ultimately updating the weight matrix of the multi-layer perceptron (MLP) by adjusting the projection layer. Similarly, methods like \cite{tian2023ftfactuality} leverage external knowledge bases to fact-check and score the quality of model responses, combining generated confidence with preference scores as rewards in a direct preference optimization (DPO) \cite{rafailov2024direct} algorithm to enhance the factuality of the model's output. Another approach \cite{lyu2024knowtuning} focuses on augmenting the knowledge perception of LLMs, integrating explicit knowledge triplet extraction and implicit multidimensional knowledge preference scoring to structurally update model parameters and strengthen the model's knowledge perception strategies.

However, the method of adjusting model parameters to inject knowledge still does not alter the black box nature of LLMs' parameters, thus, the factuality of the content they generate cannot be guaranteed. Retrieval Augmented Generation (RAG) has effectively improved the performance of language models on knowledge-intensive tasks by utilizing external knowledge sources \cite{lewis2020retrieval}. By designing specific interfaces and constructing relevant queries, LLMs gain the ability to tap into worldwide knowledge. This access helps reduce the occurrence of generated inaccuracies and the outdatedness of the knowledge encoded in their parameters \cite{nakano2021webgpt, lazaridou2022internet, semnani2023wikichat, qin2023webcpm}.To ensure that LLMs reference trustworthy sources when generating responses, the RAG technology is a worthwhile method to explore.

There has been a considerable amount of research on factual knowledge previously. These efforts extend the design principles of traditional PLM tasks, such as fact-checking or fact verification \cite{thorne2018fever, aly2021feverous, jiang2020hover}, but exclusively retain the claim part as the input for LLMs. In the knowledge-intensive domains, financial scenarios pose a more stringent test and challenge to the timeliness of world facts compared to fields like mathematics, medicine, and law.

Therefore, we propose FinFact, the first Chinese financial domain factual knowledge verification dataset. We use authoritative Chinese news media as information sources, such as Xinhua Net (http://www.news.cn/), China Youth Online (https://www.youth.cn/index.htm), and China Economic Net (http://www.ce.cn/), collecting over 200 high-quality news texts to construct over 1,500 factual question-answer pairs, ultimately forming a fact verification dataset for the financial sector.

We also introduce IDEA-FinQA, a financial question-answering system driven by LLMs. IDEA-FinQA adheres to a scheme of real-time knowledge injection and factual enhancement using external knowledge for LLMs. The system comprises three main modules: the data collector is responsible for collecting and integrating financial domain data, including data storage solutions, online and offline collection; the data querying module offers data search methods based on two types of search engines, traditional text-based indexing and popular embedding-based indexing, for multiple stages of recall and ranking; the driving force of IDEA-FinQA is four LLM-based agents, performing corresponding tasks given different prompts and contexts, including a query rewriter, intention detector, extractor and refiner, and a response generator.

Our contributions can be summarized as follows:

\begin{enumerate}
    \item We build the first Chinese financial domain factual knowledge verification dataset, utilizing authoritative Chinese news media as sources. 
    \item We construct an advanced financial question-answering system driven by LLMs. IDEA-FinQA integrates real-time financial information retrieval, text embedding search engines, and a sophisticated financial question-answering agent. 
    \item By leveraging trustworthy sources and sophisticated data retrieval and processing techniques, the IDEA-FinQA system ensures that the financial advice and information it provides are not only timely but also factually accurate.
\end{enumerate}

\section{FinFact: Financial Fact Checking Dataset}

\begin{figure}[htbp]
    \centering
\includegraphics[width=1.0\linewidth]{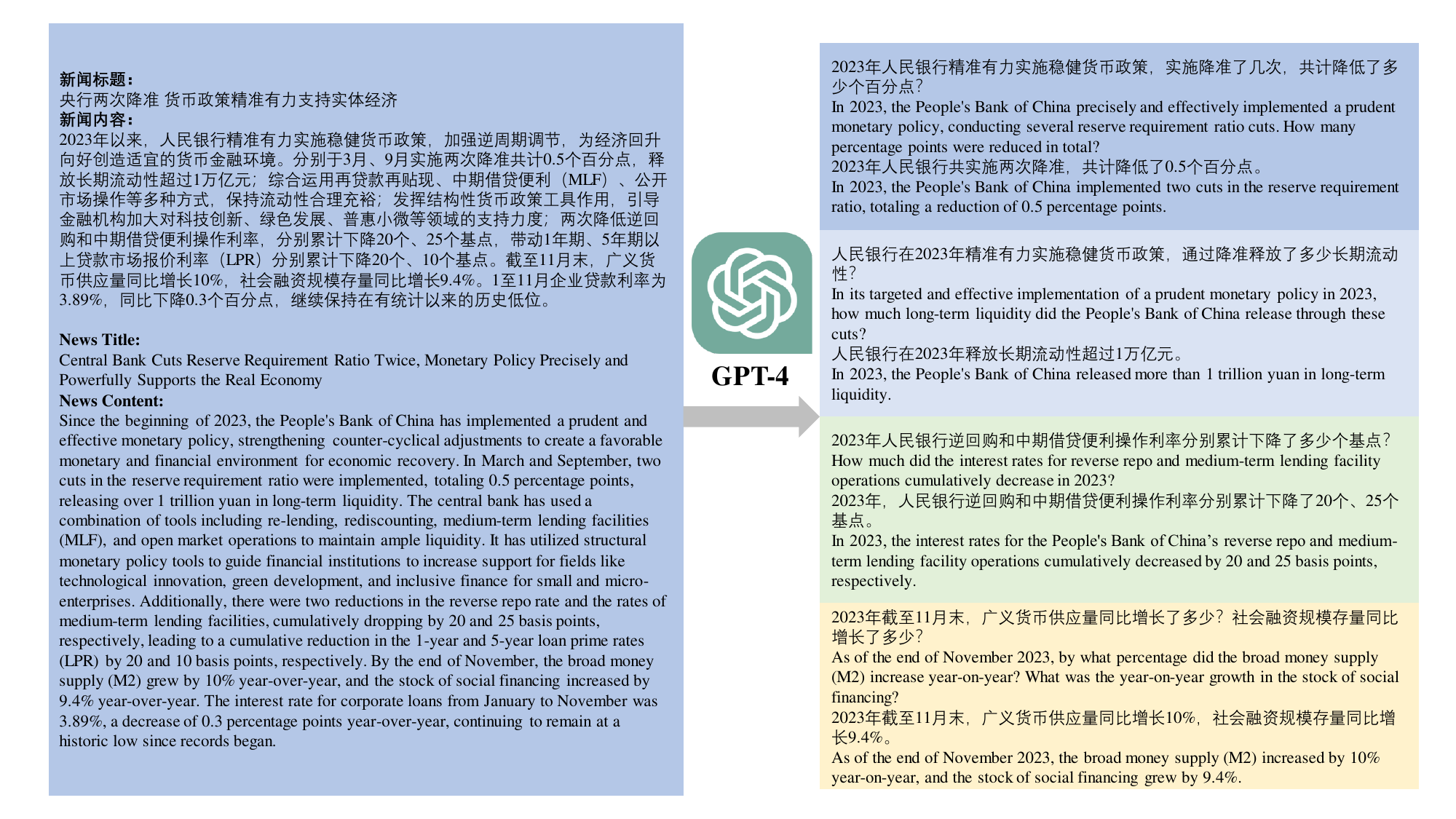}
    \caption{Use GPT-4 to generate structural questions in FinFact dataset.}
    \label{fig:finfact_structural}
\end{figure}

The creation of factual datasets, as exemplified by sources like \cite{thorne2018fever, aly2021feverous, jiang2020hover}, is a meticulous and deliberate endeavor. The development of a Chinese financial factual dataset represents a pioneering effort in this domain. It is crucial to ensure the authenticity and reliability of the information sources. To this end, we utilize reputable Chinese news outlets such as Xinhua Net (http://www.news.cn/), China Youth Online (https://www.youth.cn/index.htm), and China Economic Net (http://www.ce.cn/) as foundational resources.

Our strategy aims to encompass a broad range of topics rather than restricting the dataset to a singular domain. Consequently, we prefer to compile annual summary-type news articles rather than searching for news based on specific thematic keywords. An illustrative URL is (https://www.sohu.com/a/747923588\_118392), which links to a selection of the top ten domestic economic news stories of 2023 by China's Economic Daily. These news cover diverse themes such as macroeconomic policies, agricultural economics, real estate, China's A-shares, and industrialization, ensuring a rich content variety. Ultimately, we amassed a total of 120 financial news articles to serve as the dataset's source material. Additionally, we gathered a smaller set of 90 articles from international, technology, and sports news to further support generalization across different domains.

\begin{figure}[htbp]
    \centering
\includegraphics[width=1.0\linewidth]{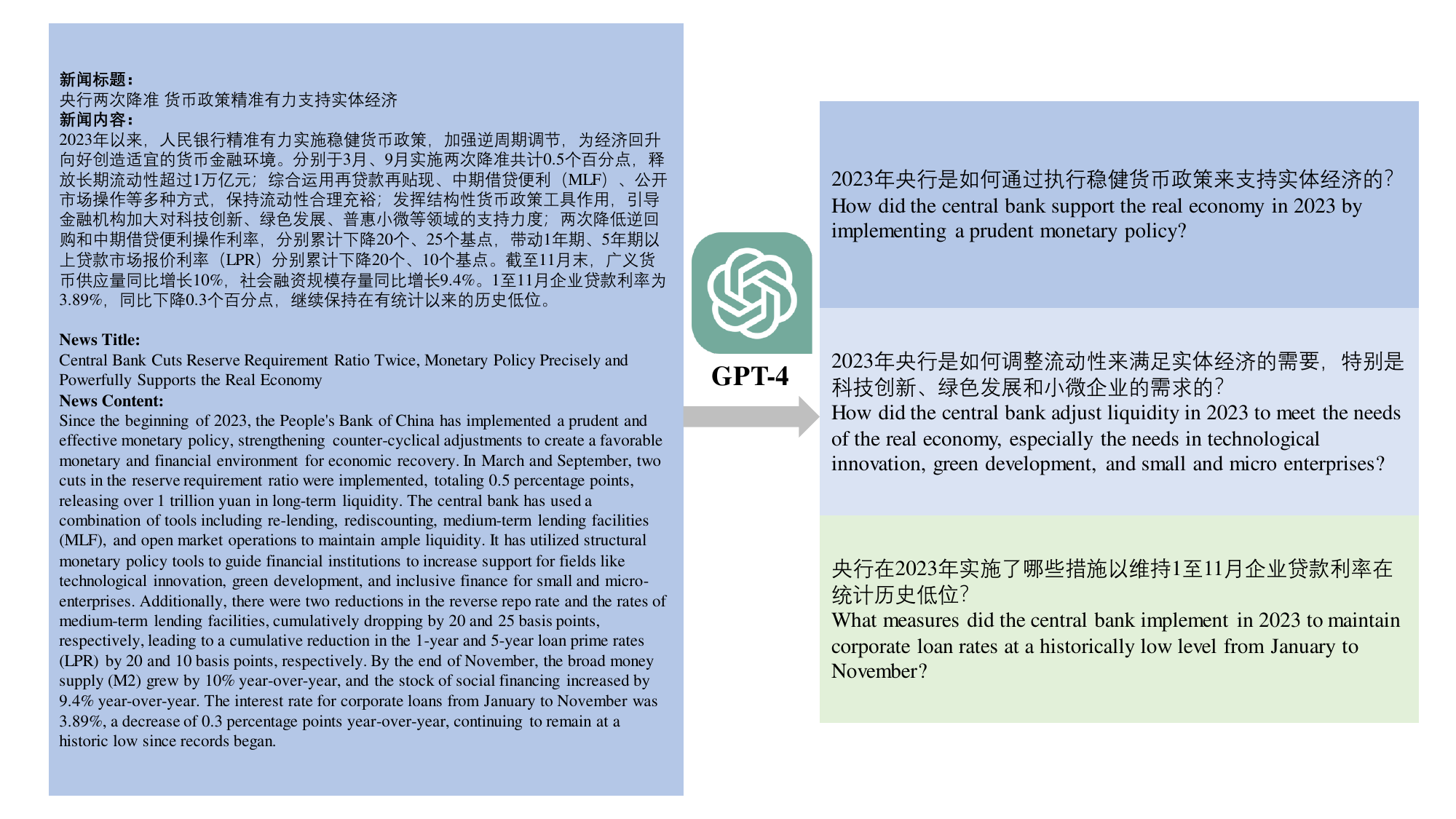}
    \caption{Use GPT-4 to generate conversational questions in FinFact dataset.}
    \label{fig:finfact_conversational}
\end{figure}

Regarding the methodology for generating inputs for LLMs, unlike the multiple-choice question format used in the previous FinKBench, the FinFact dataset interacts with LLMs through a conversational format. We utilize GPT-4 for this generation process.

In constructing the FinFact dataset, we adopted two approaches. The first approach involves structured question-and-answer sessions where we require GPT-4 to generate questions based solely on the news content, focusing exclusively on specific and objective entities. These include but are not limited to names, places, organizations, dates, and numbers, while consciously avoiding subjective opinions, attitudes, and perspectives. The corresponding answers are also generated accordingly. An example of this can be seen in Figure \ref{fig:finfact_structural}. For instance, consider the question: "In 2023, the People's Bank of China precisely and effectively implemented a prudent monetary policy, conducting several reserve requirement ratio cuts. How many percentage points were reduced in total?" This question explicitly references numerical entities, leading to a definitive answer: "In 2023, the People's Bank of China implemented two cuts in the reserve requirement ratio, totaling a reduction of 0.5 percentage points."

\begin{table}[ht]
    \centering
    \begin{tabular}{ll}
    \toprule
    \textbf{Category} & \textbf{\#Counts} \\
    \midrule
    Financial & 120 \\
    Political & 30 \\
    Technical & 30 \\
    Sports & 30 \\
    \midrule
    \textbf{Years} & \textbf{\#Counts} \\
    \midrule
    2023 & 70 \\
    2022 & 70 \\
    2021 & 70 \\
    \midrule
    \textbf{Questions} & \textbf{\#Counts} \\
    \midrule
    Structural & 877 \\
    Conversational & 637 \\
    \textit{Total} & 1514 \\
    \bottomrule
    \end{tabular}
    \caption{Statistics of FinFact datasets. The Category and the Years refer to the news sources.}
    \label{table:finfact_statistics}
\end{table}

In the construction of conversational questions, our generation strategy for GPT-4 tends to focus on questions that revolve around the exposition, attitude, viewpoint, and perspective presented in the news material concerning the event, rather than targeting specific, objective entities. Furthermore, the answers are already included within the news content itself. An example of this can be seen in Figure \ref{fig:finfact_conversational}.

\section{IDEA-FinQA: Financial QA system}

\begin{figure}[htbp]
    \centering
\includegraphics[width=1.0\linewidth]{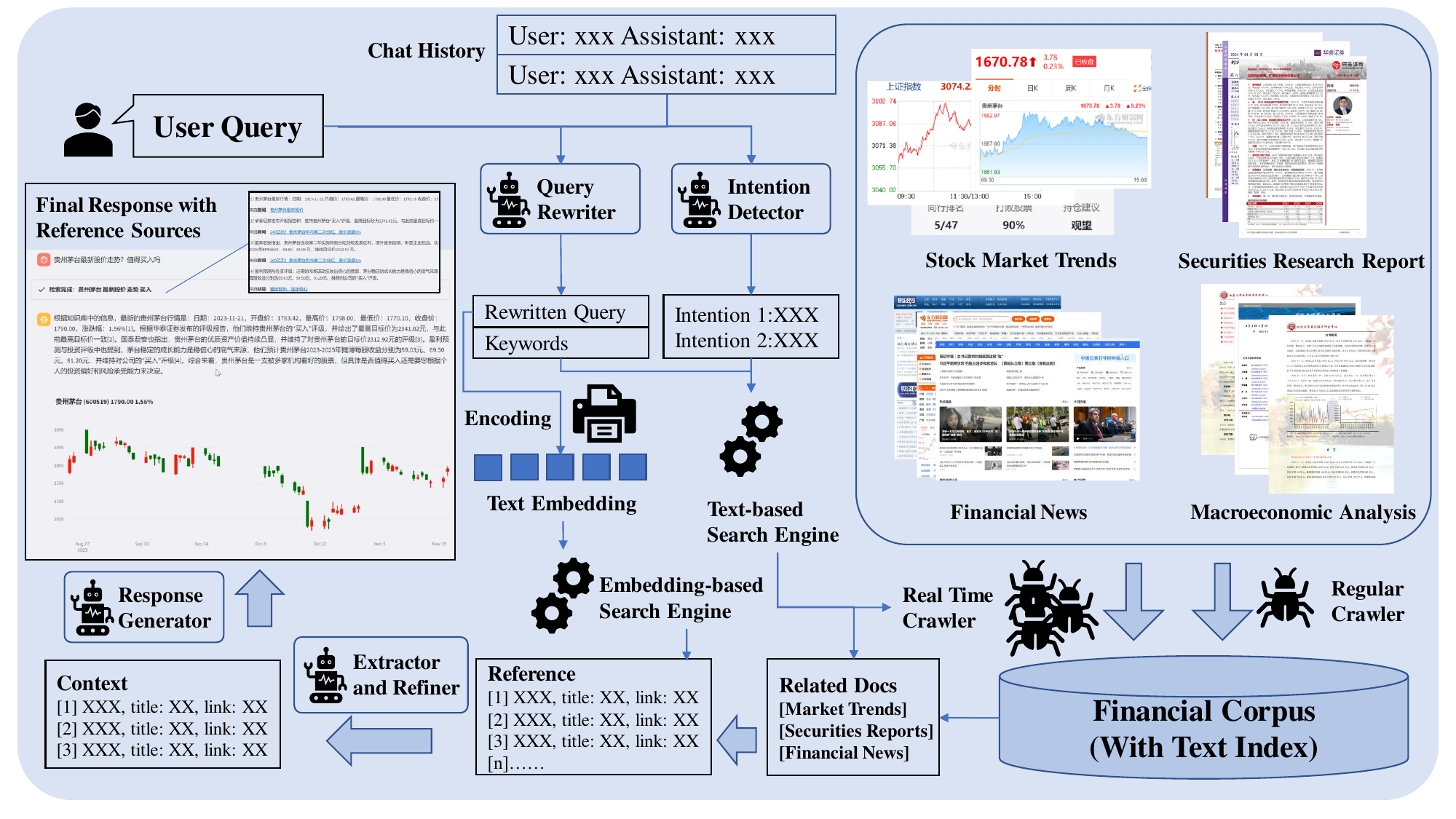}
    \caption{An overview of our proposed IDEA-FinQA system.}
    \label{fig:finqa_overview}
\end{figure}

In this section, we will dismantle the functions of each module of the IDEA-FinQA system, including the data collector, the data search engine and the LLM-driven agents. An overview of IDEA-FinQA is shown in Figure \ref{fig:finqa_overview}

\subsection{Data Collector}

\paragraph{Data Sources} Regarding data sources, we have selected four types of data, which include stock market trends, financial news, securities research reports, and macroeconomic analyses—all of which are textual in nature. Stock market data is sourced from the East Money website (https://www.eastmoney.com/). By specifying a stock name, such as Kweichow Moutai, and a particular time period, such as one week, we can gather data suitable for creating candlestick charts reflecting stock movements, including daily opening, closing, highest, and lowest prices. Additionally, technical stock analyses, including support levels, resistance levels, valuations, and trading recommendations, are also obtained from the market center on the same website. The sources for financial news include two financial websites, East Money and Finance Website (https://www.caijing.com.cn/), and a general news website, Tencent (https://www.qq.com/). All provide high-quality news written and verified by professionals. The news list can be accessed by entering search keywords into the developed interface, and clicking any link retrieves the text of the news. Finally, securities research reports and macroeconomic analyses are sourced daily from the research report center on East Money, offering high-quality reports from professional institutions covering individual stocks, industries, macroeconomics, and strategies. However, this site does not offer a text search tool.

\paragraph{Crawler} The crawlers for the aforementioned data are categorized into two types. One is for long-term collection and periodic updates, such as the securities research report and macroeconomic crawlers. We collect the latest research content daily, saving the report titles, summaries, and PDF links in a structured format in a local database. The second type of crawler is real-time, including for stock market trends and financial news, which requires specific search terms to navigate to the websites and use the search engines to retrieve and return the most relevant sorted news list.

\subsection{Data Search Engine}

\paragraph{Text-based Index} The text index is primarily used for keyword retrieval in the local database, such as for research reports. We employ traditional text indexing methods, segmenting the titles and summaries of each report into Chinese words, combining them with company names, assigning weights for text matching scores, and prioritizing the most recent articles. Ultimately, we have built a search engine for our research report database locally. This is similar to the search engine functionalities inherent in news websites, primarily used for preliminary filtering of texts under given query statements as candidates.

\paragraph{Embedding-based Index} The embedding index is mainly used for retrieval based on text similarity, returning texts by paragraph. Given that our data sources naturally include paragraph segmentation, we split candidate texts using line breaks to obtain multiple paragraphs for matching. Cosine similarity is used to calculate the similarity scores between the query statements and the candidate paragraphs, ultimately returning texts with high similarity scores as credible and highly relevant sources of knowledge.

\subsection{LLM-driven Agents}

\paragraph{Query Rewriter} This agent receives user queries and dialogue history, calling on a LLM to rewrite them and extract suitable keywords for search. Initially, the user's query may be related to the dialogue history. For example, if "technical advantages of the Tongyi Qwen large model" were mentioned in a previous conversation, and the current query is "How does it compare to the Wenxinyiyen large model?", it should be rewritten as "Comparison of the technical advantages between the Wenxinyiyen and Tongyi Qwen large models." Additionally, the user's query may contain redundant information. For instance, if the query is "Please introduce the configurations of the newly launched XiaoMi su7 car," the rewriter will distill this to "XiaoMi su7 car configurations" as a phrase suitable for direct input into a text search engine.

\paragraph{Intention Detector} This agent is responsible for identifying the user's underlying search intent to select appropriate data sources. For example, if the query is "Is Kweichow Moutai worth holding?", this indicates a need to retrieve data related to the stock trends of Kweichow Moutai, combined with recent research reports. Additionally, news and stock market opinions may be consulted as part of a custom temporary knowledge base.

\paragraph{Extractor and Refiner} This agent is used for extracting and refining knowledge from the retrieved database. Since the context length is typically extensive to ensure coverage of the content needed to answer the query, the scale of external knowledge required varies by question. For instance, answering "How many stocks are there in the A-share market currently?" may need only one or two pieces of external knowledge, whereas "Analyzing the investment value of BYD in 2024" would require a more comprehensive knowledge base. This agent also helps in refining the database to minimize style discrepancies among different entries, ensuring uniformity.

\paragraph{Response Generator} This agent generates the final response to the user's queries. Given a knowledge base in the context, it is encouraged to extract, summarize key knowledge points, and generate responses. It also needs to specify citation sources using indices like “[1][3]”. By combining the indices of retrieved knowledge base pages, this agent ensures that all information produced has corresponding referenced sources, maintaining the factuality and authority of the responses.

\section{Experimental Settings}

Our IDEA-FinQA system utilizes the open-source model Qwen1.5-14B-Chat \cite{bai2023qwen} to power all its agents. To test the fact-based question-answering capability of IDEA-FinQA on the FinFact dataset, we selected the vanilla Qwen1.5-14B-Chat, Yi-34B-Chat \cite{young2024yi}—one of the strongest current Chinese models, and the globally popular GPT series. Since the GPT series is only available through APIs, we chose gpt-4-turbo-preview and gpt-3.5-turbo for testing, with the model knowledge updated up to December 2023 at the time of testing. For each question in FinFact, we collect and save the complete text returned by the LLMs. Subsequently, GPT-4 is used as a judge to assess the quality of the LLM outputs. For structural questions, standard answers generated during the dataset creation phase are also input to evaluate factuality, whereas for conversational questions, original news texts are input to similarly assess factuality. To objectively evaluate the generated response quality, we also incorporate "relevant" and "informational" as additional dimensions for assessment.

\section{Result \& Analysis}

\begin{figure}[htbp]
    \centering
\includegraphics[width=0.7\linewidth]{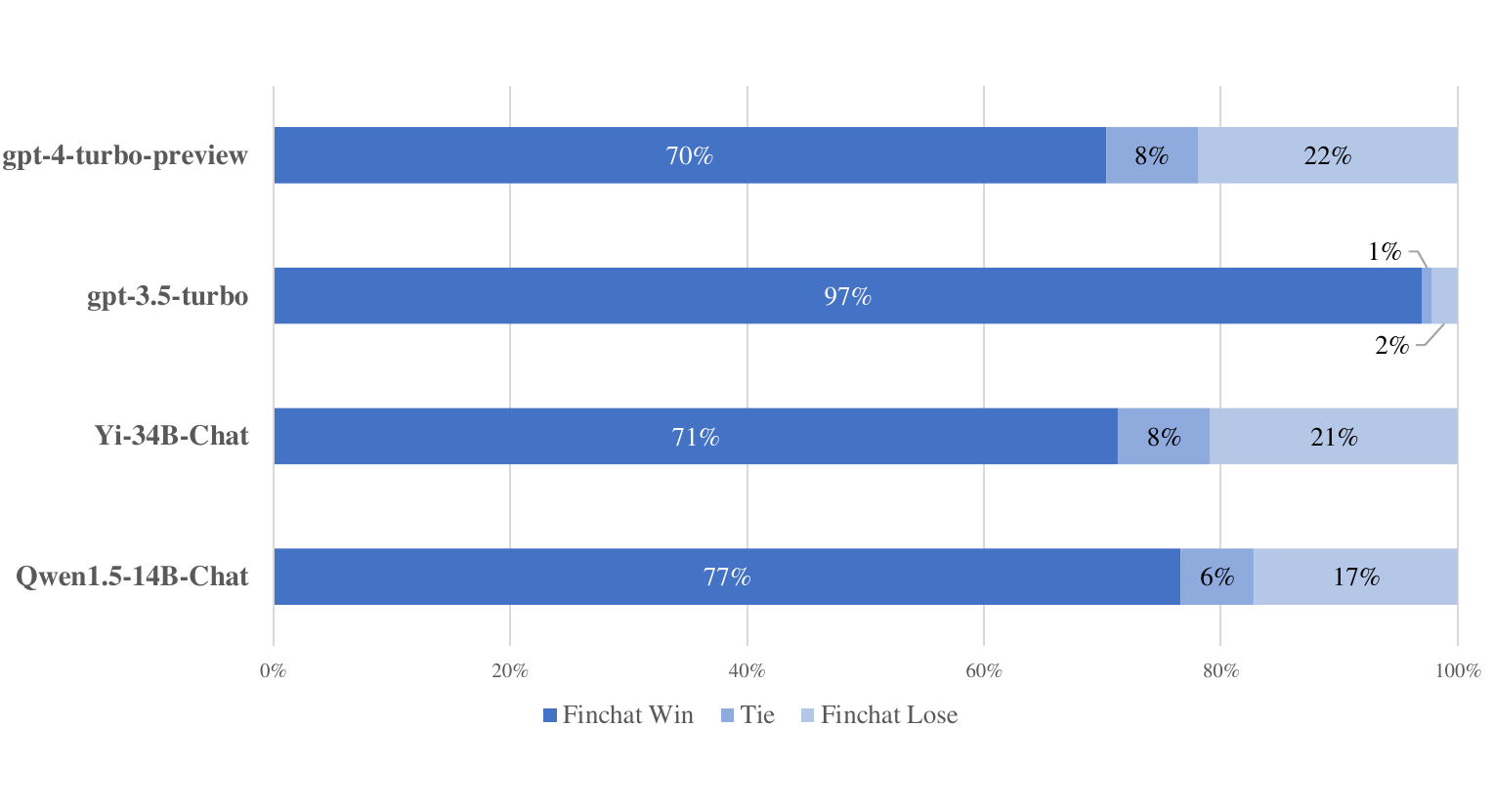}
    \caption{Evaluation of "factual" using GPT-4 as judge.}
    \label{fig:pk_factual}
\end{figure}

IDEA-FinQA demonstrates a distinct advantage in fact-based question-answering compared to other models. Across the three dimensions—factual, relevant, and informational—IDEA-FinQA leads all other models. Particularly in the factual dimension, IDEA-FinQA outperforms other models with a winning rate of 70\%. In assessments of text generation quality, including relevance and informational content, IDEA-FinQA also exhibits strong performance.

\begin{figure}[htbp] 
    \centering
\includegraphics[width=0.7\linewidth]{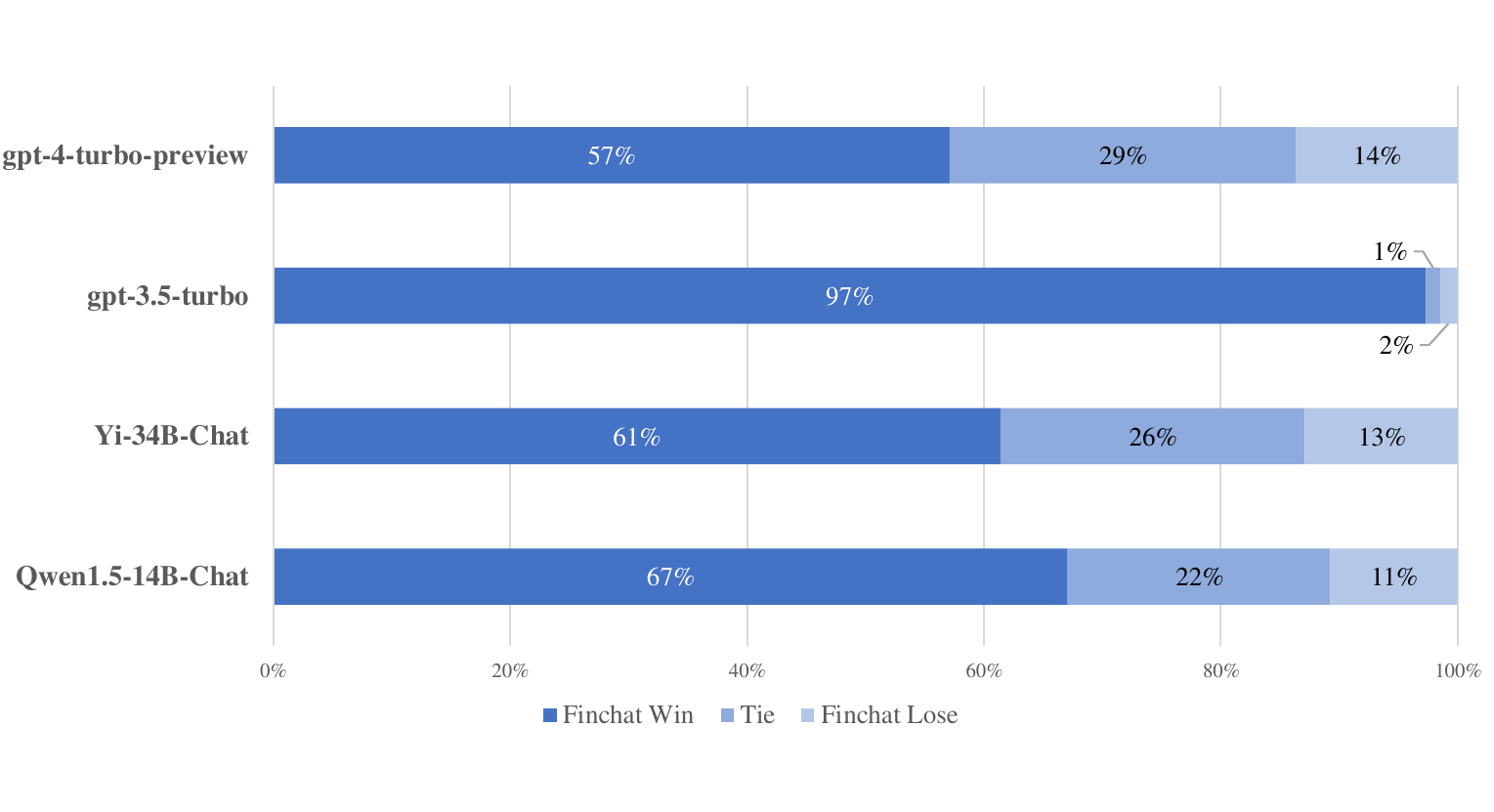}
    \caption{Evaluation of "relevant" using GPT-4 as judge.}
    \label{fig:pk_relevant}
\end{figure}

\begin{figure}[htbp]
    \centering
\includegraphics[width=0.7\linewidth]{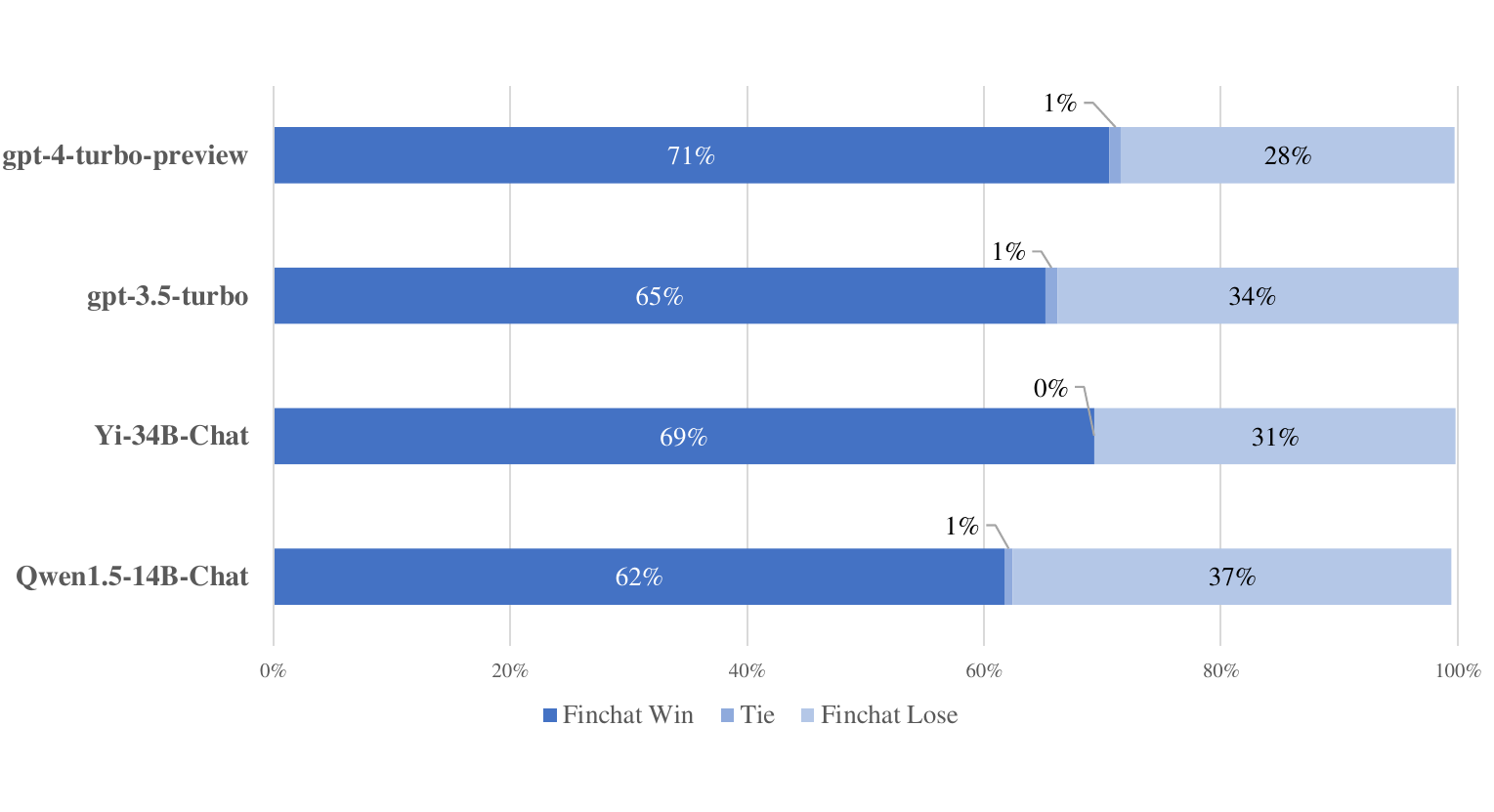}
    \caption{Evaluation of "informational" using GPT-4 as judge.}
    \label{fig:pk_informational}
\end{figure}

\section{Summary}

We first introduce FinFact, the first Chinese financial domain factual knowledge verification dataset. Subsequently, we present IDEA-FinQA, a financial question-answering system driven by LLMs. IDEA-FinQA adheres to a scheme of real-time knowledge injection and factual enhancement using external knowledge for LLMs. Our experiments demonstrate that IDEA-FinQA surpasses the majority of models in factual question-answering, even when the facts are from different years.
\newpage

\chapter{Conclusion}
\label{chp_conclusion}
This thesis discusses how factual knowledge can be applied to enhance the construction of a trustworthy LLM for the financial sector. Firstly, we established FinKBench, the first bilingual (Chinese and English) benchmark utilizing expert-level financial certification exam questions to assess the financial knowledge mastery of LLMs. Secondly, we proposed the FinKER framework to explore how financial knowledge can be effectively injected into LLMs and the different paradigms of injection. Additionally, addressing the challenge of financial fact-based question answering, we introduced the FinQA system, which combines external knowledge bases and LLM-driven agents to provide authoritative and trustworthy citation sources. This system demonstrated impressive performance on the factual dataset FinFact.

In Chapter \ref{chp_finkbench}, we introduce FinKBench, an evaluation benchmark for financial knowledge in LLM, utilizing questions from two globally renowned and authoritative financial professional exams as the primary sources for assessment. The questions, encompassing both Chinese and English languages, four types of question formats, and spanning sixteen financial disciplines, are designed to evaluate LLMs' capabilities in directly addressing exam questions relevant to the finance sector comprehensively. Additionally, we provide a modular evaluation suite that can incorporate external datasets, allowing for flexible customization of evaluation modes and interfaces with various LLMs, thus offering adaptability and scalability to the evaluation framework. 

In Chapter \ref{chp_finker}, we introduce FinKER, which is a Financial Knowledge Enhancement framework. FinKER is designed to facilitate the rapid adaptation of general LLMs to the financial domain without incurring the high costs associated with external pre-training. This framework is supported by a meticulously cleaned and constructed comprehensive database of Chinese financial exam questions, which incorporates support embedding similarity retrieval. FinKER underpins the development of a retrieval-based few-shot learning method for real-time context-level knowledge injection, termed soft-injecting paradigm of knowledge. Additionally, we have developed a high-quality set of financial knowledge instructions for fine-tuning any general LLM, referred to as hard-injecting paradigm of knowledge. Empirical evidence demonstrates that FinKER significantly enhances the expert capabilities of LLMs within the financial domain, notably improving their performance on the FinKBench, especially in the segment pertaining to Chinese exam questions like CPA.

In Chapter \ref{chp_finqa}, we introduce FinQA, a financial question-answering system driven by LLMs. FinQA adheres to a scheme of real-time knowledge injection and factual enhancement using external knowledge for LLMs. The system comprises three main modules: the data collector is responsible for collecting and integrating financial domain data, including data storage solutions, online and offline collection; the data querying module offers data search methods based on two types of search engines, traditional text-based indexing and popular embedding-based indexing, for multiple stages of recall and ranking; the driving force of FinQA is four LLM-based agents, performing corresponding tasks given different prompts and contexts, including a query rewriter, intention detector, extractor and refiner, and a response generator.

The primary contribution of this thesis is the exploration of how to build trustworthy LLMs in the specific domain of finance by employing methods enhanced with factual knowledge. Artificial intelligence has already begun to enable the finance industry, and the robust performance of LLMs further fuels professionals' expectations for automated agents. By proposing a financial knowledge benchmark, knowledge injection paradigms, and an externally enhanced dialog system, this work makes significant contributions worthy of reference in the field.
\newpage

\addcontentsline{toc}{chapter}{Bibliography}
\bibliographystyle{IEEEtran}
\bibliography{references}
\newpage



\end{document}